\pdfoutput=1
\documentclass[]{amap}

\usepackage{booktabs}
\usepackage{multirow}
\usepackage{graphicx}
\usepackage{array}
\usepackage{siunitx}
\usepackage[table]{xcolor}
\usepackage{makecell}
\usepackage{framed}
\usepackage{hyperref}
\usepackage{amsmath}
\usepackage{amsfonts}
\usepackage{placeins}
\usepackage{longtable}
\usepackage{hhline}
\usepackage{fancyvrb}
\usepackage{float}
\usepackage{fvextra}
\usepackage{multicol}
\usepackage{cleveref}
\usepackage{tablefootnote}
\usepackage{threeparttable}
\usepackage{tabularx}
\usepackage{mdframed}
\usepackage[usestackEOL]{stackengine}
\usepackage[numbers]{natbib}
\usepackage{enumitem}
\usepackage{adjustbox}
\usepackage{arydshln}
\usepackage{pifont}
\usepackage[dvipsnames]{xcolor}

\renewcommand{\paragraph}[1]{\noindent\textbf{#1.}\hspace*{1em}}
\setlist[itemize]{leftmargin=15pt}

\title{PhysBrain 1.5: From Vision-Language Models to Physical Foundation Models}

\author{DeepCybo Team}
\vspace{-10pt}

\projectpage{https://deepcybo-physai.github.io/PhysBrain-1.5}
\checkdata[Huggingface]{\url{https://huggingface.co/collections/DeepCybo/physbrain-15}}
\checkdata[Evaluation Kit]{\url{https://github.com/DeepCybo-PhysAI/PhysBrainEvalKit}}

\vspace{-10pt}
\abstract{We present PhysBrain 1.5, a unified model for understanding physical environments, generating actions, and predicting future states. Motivated by the physical loop of observation, interaction, and environmental change, we bring these capabilities into a common learning framework. Starting from a general vision--language model, we encode language responses, end-effector motion, and dense visual targets as discrete sequences and jointly optimize them with autoregressive next-token prediction. Pre-training draws its embodied supervision entirely from human interaction videos, using task-centered episodes to pair semantic and spatial context with recovered motion and subsequent observations. We then adapt the model through supervised fine-tuning on a mixture of human demonstrations, robot trajectories, and simulated experience. Across 28 embodied understanding benchmarks, our 8B model achieves an average score of 72.5, setting a new open-source state of the art and performing on par with leading proprietary models such as GPT-6-Astra and Gemini 3.6 Flash. It achieves the best open-source results on 14 benchmarks while retaining general multimodal capabilities. Beyond these understanding evaluations, qualitative examples show the model's ability to produce end-effector trajectories and predict future scenes through spatially aligned RGB, depth, and robot-mask outputs.
}
\vspace{-10pt}

\begin{document}
\maketitle

\begin{figure}[H]
\centering
\vspace{-20pt} 
\makebox[\linewidth][c]{\includegraphics[width=1.025\linewidth]{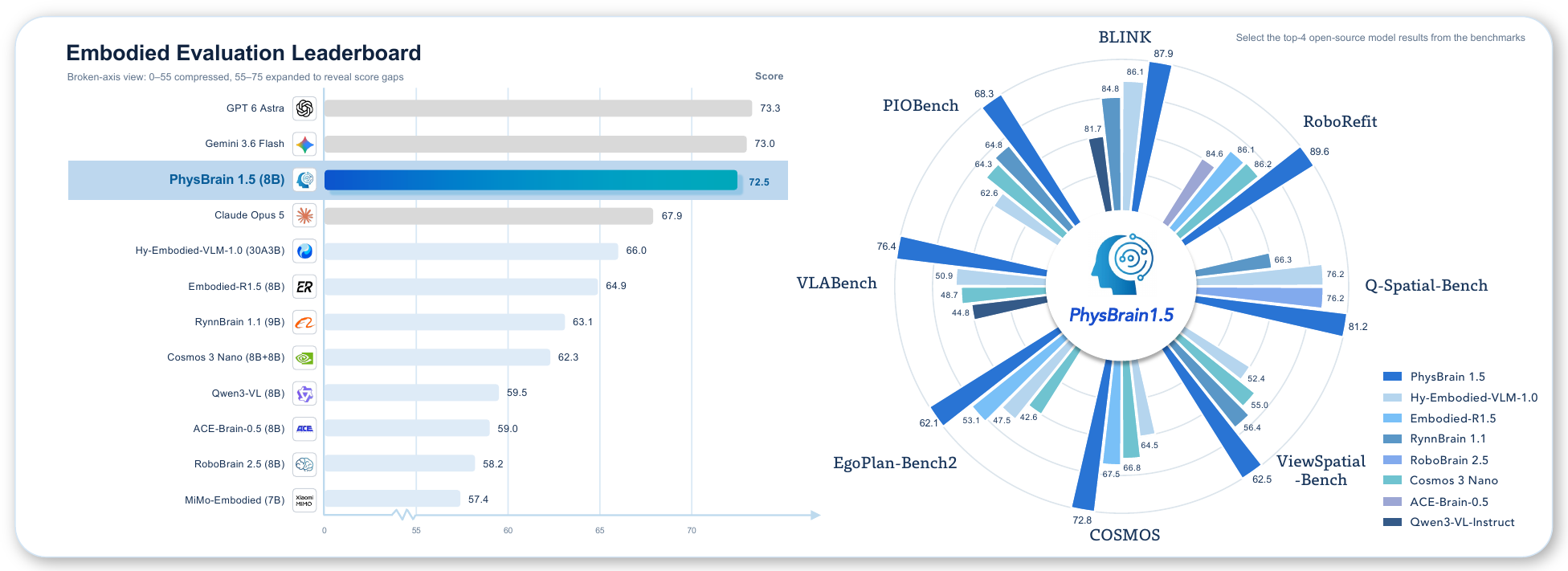}}
\caption{\textbf{Leading overall performance on the report's embodied benchmark suite.}
PhysBrain~1.5 achieves an overall score of \textbf{72.5} with 8B parameters, ranking first among all open-sourced models and delivering performance on par with leading proprietary systems. }
\vspace{-6pt} 
\label{fig:overall}
\end{figure}

\clearpage
\tableofcontents
\clearpage

\section{Introduction}
\label{sec:intro}

Physical intelligence depends on continuous interaction between an agent and its environment. We describe this interaction as the \emph{physical loop}: the agent observes its surroundings, interprets spatial relationships and task goals, anticipates action outcomes, and acts on the environment. The changed world provides new observations that inform its understanding and subsequent behavior. A physical foundation model should provide reusable capabilities for understanding, acting, and predicting within this loop.

The capabilities needed for this loop have developed across general foundation models and specialized systems. General vision--language models (VLMs) provide visual understanding, semantic knowledge, and instruction following~\citep{Bai2025Qwen3Vl,Zhu2025Internvl3Exploring,Deitke2025MolmoAnd}. Specialized models address spatial reasoning~\citep{Song2025RobospatialTeaching,Ray2025SatDynamic,Ma2025SpatialreasonerTowards,Yang2026CambrianS}, grounding and affordance~\citep{Yuan2024RobopointA,Zhou2025RoboreferTowards,Clark2026MolmopointBetter}, planning~\citep{Sermanet2024RobovqaMultimodal,Chen2023EgoplanBench}, and execution assessment~\citep{Lee2026RoborewardGeneral,Liang2026RobometerScaling}. Vision--language--action models (VLAs), including the $\pi$ series, GR00T, and Qwen-VLA, learn robot behavior~\citep{Black2024_0A,PhysicalIntelligence2025_05,PhysicalIntelligence2026_07,NVIDIA2025Gr00tN1,Wang2026QwenVla}. DreamZero and Qwen-RobotWorld incorporate predictive world modeling~\citep{Ye2026WorldAction,Zhang2026QwenRobotworld}. At the system level, SayCan~\citep{Ahn2022DoAs}, Code as Policies~\citep{Liang2023CodeAs}, and VoxPoser~\citep{Huang2023VoxposerComposable} connect language reasoning with robot capabilities through skills, programs, or spatial control representations. These approaches advance physical interaction by training specialized models and integrating existing capabilities into robotic systems.

Recent embodied foundation models bring several of these capabilities together. HY-Embodied, RynnBrain~1.1, and MiMo-Embodied broaden spatial understanding, interaction reasoning, and planning~\citep{TencentRoboticsX2026HyEmbodied,Wang2026HyEmbodied,Li2026Rynnbrain1,Hao2025MimoEmbodied}. Other models integrate grounding, decisions, and execution-related capabilities~\citep{Tan2026Robobrain2,Yuan2026EmbodiedR1,ACEBrainTeam2026AceBrain}. UniVLA and RynnVLA-002 explore shared learning of understanding, action generation, and visual prediction~\citep{Wang2025UnifiedVision,Cen2025Rynnvla002}, while Cosmos~3 develops a broader framework for multimodal understanding and generation in physical environments~\citep{NVIDIA2026Cosmos3}. Building on these efforts, we study how broad embodied understanding can be learned jointly with end-effector motion and multimodal future states.

Joint learning must account for differences in supervision. Understanding tasks produce language, spatial coordinates, and interaction targets. Action generation requires temporally structured motion representations, while future-state prediction involves dense visual outputs. Training data also differ in embodiment, annotation conventions, and temporal granularity. We seek a shared learning formulation that accommodates these differences while supporting physical generation alongside broad embodied understanding.

We present \textbf{PhysBrain~1.5}, which \textbf{unifies embodied understanding, action generation, and future-state prediction} through discrete tokens. We extend the language vocabulary with action and visual tokens. ActionPiece~\cite{DeepCyboTeam2026ActionpieceRethinking} encodes end-effector trajectories, while visual tokens represent future RGB images, depth maps, and robot masks. These outputs share an autoregressive backbone and output head, allowing semantic, motion, and future-state supervision to update the same parameters through a common next-token prediction objective, with task-specific loss masks selecting the target tokens to supervise for each output type.

Human interaction experience provides the data foundation for this approach. Our earlier work, PhysBrain~1.0~\citep{Lian2026Physbrain1}, converts human egocentric video into structured physical-commonsense supervision and transfers the learned priors to robot policies. PhysBrain~1.5 extends this understanding-first approach to joint learning across the physical loop, with all embodied pre-training supervision derived from human interaction experience. We organize egocentric and synchronized ego--exocentric recordings, together with panoramic video from PhysBrain-Ego360~\citep{DeepCyboTeamndEgo360A}, into task-centered episodes. These episodes couple semantic and spatial annotations, recovered human motion~\citep{Lin2026HumanAs}, and subsequent visual states within the shared training formulation. Supervised fine-tuning then combines human demonstrations, real-robot trajectories, and simulated interactions to extend the learned capabilities to diverse embodiments and tasks.

We evaluate PhysBrain~1.5 on 28 benchmarks spanning perception, spatial and multi-view understanding, embodied reasoning and planning, grounding and affordance, and visual trajectory reasoning. As shown in Figure~\ref{fig:overall}, our 8B model achieves an average score of \textbf{72.5}, setting a new \textbf{open-source state of the art} and closely approaching the strongest proprietary models, GPT-6-Astra (73.3) and Gemini~3.6~Flash (73.0). It ranks first on \textbf{14 benchmarks} and second on \textbf{10} among open-source models. Qualitative results further demonstrate end-effector trajectory generation and future-frame prediction.

\section{Model Architecture}
\label{sec:model}

\subsection{Overview}

\begin{figure}[!t]
    \centering
    \includegraphics[width=1.0\linewidth]{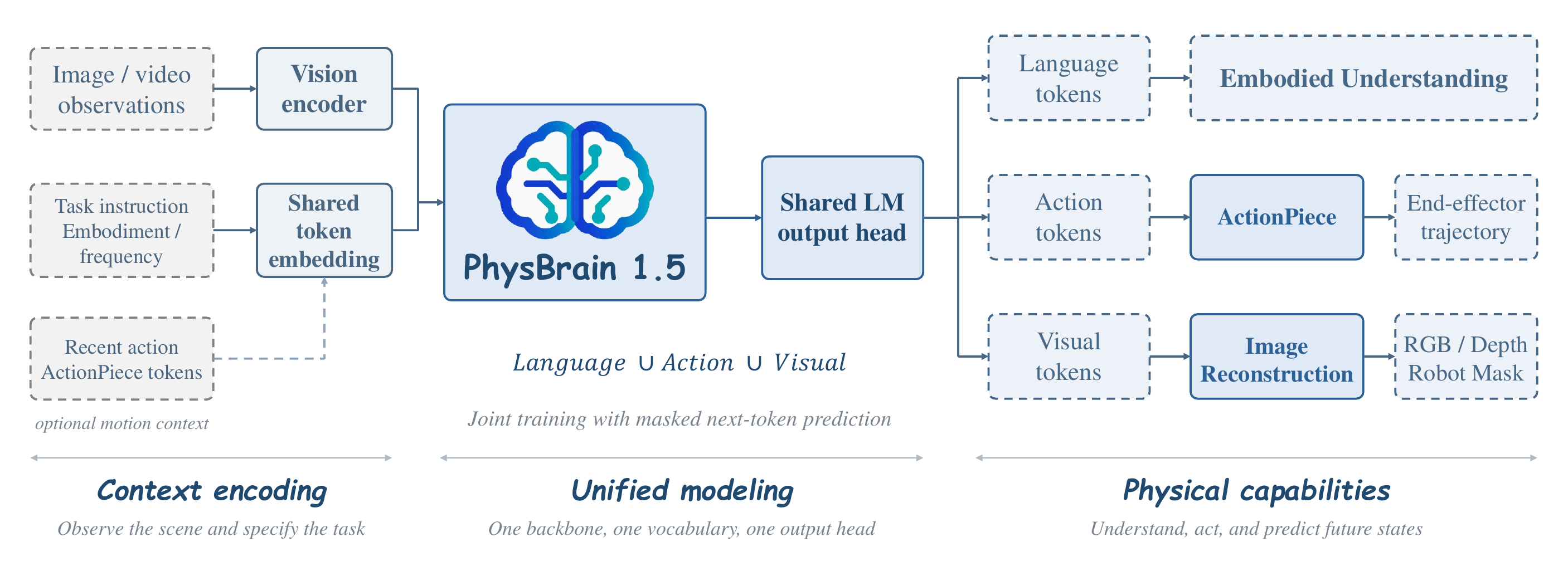}
    \caption{\textbf{PhysBrain~1.5 model architecture.}
    Built on a pretrained vision--language model, PhysBrain~1.5 jointly learns embodied understanding, action generation, and visual-state prediction through a shared backbone, token embedding, and output head over a unified language--action--visual vocabulary. Embodied understanding covers foundational visual-spatial perception, spatial and multi-view understanding, embodied cognition and reasoning, spatial pointing and affordance understanding, and visual trajectory reasoning. Action tokens are decoded by ActionPiece into end-effector trajectories, with optional action history providing motion context; visual tokens are decoded by a Vision decoder into future RGB images, depth maps, and robot masks. The three task families provide complementary supervision under a shared autoregressive training objective.}
    \label{fig:model_architecture}
\end{figure}

PhysBrain~1.5 is built from the Qwen3-VL Instruct family and extends a
pretrained vision--language model into a unified model of physical perception,
interaction, and state transition. As illustrated in Figure~\ref{fig:model_architecture},
the central design is to express language,
end-effector motion, and visual-state targets as discrete sequences and learn
them through the same autoregressive interface. The unification concerns the generated
representations, which share the language-model backbone, token embedding, and
output head. Specifically, we augment the original language vocabulary with
dedicated sets of action tokens and visual tokens:
\begin{equation}
    \mathcal{V}=\mathcal{V}_{\mathrm{lang}}\cup\mathcal{V}_{\mathrm{act}}\cup\mathcal{V}_{\mathrm{vis}}.
    \label{eq:unified_vocab}
\end{equation}
The resulting vocabulary supports language, action, and visual-state generation
through a shared embedding table and language-model output head. This
formulation enables joint training of understanding and generation: perceptual
supervision provides context for predicting interactions and their consequences,
while action and state-transition supervision introduces physical priors that
support spatial grounding and trajectory reasoning.

Let $x$ contain the visual input, instruction, and any task-dependent context,
and let $y$ denote the target sequence. Across task families, the model is
optimized with masked next-token prediction:
\begin{equation}
    p_{\theta}(y\mid x)=\prod_{i=1}^{|y|}p_{\theta}(y_i\mid x,y_{<i}),\qquad
    \mathcal{L}_{\mathrm{AR}}=-\sum_{i=1}^{|y|}m_i\log p_{\theta}(y_i\mid x,y_{<i}),
    \label{eq:unified_ar}
\end{equation}
where $m_i$ selects supervised target tokens. Task formatting, modality
delimiters, and the loss mask specify whether the model should answer in
language, produce an action trajectory, or generate a visual-state sequence;
the underlying prediction objective remains unchanged.

\subsection{Embodied Understanding as Physical Perception}

Embodied understanding is not implemented as a separate perception head.
Instead, PhysBrain~1.5 retains the general image and video understanding
interface of the base VLM and enriches it through embodied supervision. This
supervision covers physical and spatial perception, multi-view understanding,
embodied reasoning and planning, object and part grounding, affordance, and
visual trajectory reasoning. Depending on the task, the same autoregressive
decoder produces natural-language answers or structured spatial targets.
Consequently, semantic recognition, spatial localization, and temporal
reasoning remain available through a common interface and can provide the
scene-level evidence needed by action and visual-state generation.

\subsection{Embodied Action Generation}
\label{sec:embodied_action_generation}

We convert human motion into robot action representations through Human-as-Humanoid~\citep{Lin2026HumanAs}, retaining the positions and orientations of both wrists and excluding finger motions. Human-derived and robot action data then share the same action representation and tokenizer: we model each action as a short end-effector trajectory. For each human or robot wrist, the model predicts an action chunk of length $H=16$, corresponding to the next 16 action steps, relative to the wrist pose at the current observation. We represent relative orientation using a continuous 6D representation. For a rotation matrix $R$, we define

\begin{equation}
    \rho_6(R) = \operatorname{vec}\!\left(R_{[:,1:2]}\right) \in \mathbb{R}^{6},
    \label{eq:rotation_6d}
\end{equation}

where $R_{[:,1:2]}$ denotes the first two columns of $R$. The action target is then defined as:

\begin{equation}
 a_{t,k}=\left[p_{t+k}-p_t,\rho_6\!\left(R_t^{\top}R_{t+k}\right),g_{t+k}\right]\in\mathbb{R}^{10},\quad k=1,\ldots,H.
 \label{eq:eef_action}
\end{equation}

The first three dimensions are relative translation, the next six encode the relative rotation, and $g\in[0,1]$ is absolute gripper closure, with zero denoting open and one denoting closed. All targets share the same anchor $(p_t,R_t)$ and are not accumulated stepwise. Physical units are standardized and translation uses corpus-level normalization; each source's native coordinate axes are retained.

We discretize these trajectories with an ActionPiece tokenizer $\mathcal{T}_{\mathrm{AP}}$~\citep{DeepCyboTeam2026ActionpieceRethinking}, trained on 28.7 million 16-step end-effector trajectory segments corresponding to 459.2 million action timesteps. The tokenizer provides a 512-token action vocabulary as an external discretization interface. In the configuration used by PhysBrain~1.5, it encodes a wrist trajectory into a sequence of 32 discrete action tokens. The same ActionPiece vocabulary and codebook are used for left wrist and right wrist trajectories.

\paragraph{Homogeneous action continuation}
Although action trajectories from different sources are mapped to the same 10D end-effector representation and ActionPiece vocabulary, we do not impose a single globally canonicalized coordinate frame or motion convention across all embodiments. Different datasets use substantially different robot, camera, and control conventions, making reliable global canonicalization costly and difficult to scale across embodiments. Moreover, some sources lack complete calibration metadata or global-frame information, so the corresponding transformations cannot be recovered without additional assumptions. Rather than introducing potentially unreliable conversions, we preserve each source's native coordinate axes. The same nominal action dimensions may therefore exhibit source-specific local axes, motion scales, control frequencies, and temporal patterns.

To account for this heterogeneity, we use the action immediately preceding the current observation as a local motion context. Together with the current visual observation, recent action history provides an implicit system-identification signal, allowing the model to infer the local action convention, motion trend, and short-term response pattern of the current embodiment or data source. Rather than requiring every source to be transformed into a globally canonical action frame, the model can use this input--motion context to continue the ongoing trajectory naturally.

The preceding and target action chunks are encoded with the same ActionPiece tokenizer and action vocabulary:
\begin{equation}
 z^{-}=\mathcal{T}_{\mathrm{AP}}(A^{-}),\qquad z^{+}=\mathcal{T}_{\mathrm{AP}}(A^{+}),\qquad
 p_{\theta}\!\left(z^{+}\mid I,u,e,f,z^{-}\right),
 \label{eq:action_continuation}
\end{equation}
where $A^{-}$ and $A^{+}$ denote the preceding and immediately following 16-step action chunks, respectively, and $z^{-}$ and $z^{+}$ are their corresponding discrete action-token sequences. Here, $I$ is the current observation, $u$ the task instruction, $e$ the robot embodiment, and $f$ the action frequency; $z^{-}$ is optional. This formulation turns action prediction into conditional trajectory continuation rather than isolated chunk regression. Action history provides a local motion prior that helps preserve continuity across action-chunk boundaries and adapt to embodiment-specific motion direction, scale, and rhythm. 

\subsection{Visual-Foundation Generation}
\label{sec:visual_foundation_generation}

We model the future physical state as a spatially aligned combination of an RGB
image, a depth map, and a robot mask. Given the current RGB observation
and task instruction, PhysBrain 1.5 predicts the future physical state $S_{t+\Delta}$:
\begin{equation}
    S_{t+\Delta}
    =
    \left(
        X^{\mathrm{rgb}}_{t+\Delta},
        X^{\mathrm{depth}}_{t+\Delta},
        X^{\mathrm{mask}}_{t+\Delta}
    \right),
    \label{eq:future_physical_state}
\end{equation}
The three targets correspond to the same timestamp and image coordinate
system. For datasets with different frame rates, the target-frame offset is determined from the specified prediction interval and each source’s native frame rate.

\paragraph{Shared discrete visual representation}
We tokenize all three modalities with a shared VQ-VAE at a target resolution
of $128\times128$. RGB images are normalized to the tokenizer's input range;
 depth maps are linearly mapped to $[-1,1]$, and  robot masks are mapped
to the same range. Single-channel depth maps and masks are replicated across
three channels before tokenization. For each modality
$m\in\{\mathrm{rgb},\mathrm{depth},\mathrm{mask}\}$, the tokenizer produces
\begin{equation}
    Q^{m}_{t+\Delta}
    =
    \mathcal{T}_{\mathrm{VQ}}
    \left(X^{m}_{t+\Delta}\right)
    \in \{1,\ldots,K\}^{16\times16},
    \label{eq:visual_state_tokenization}
\end{equation}
where $K=16{,}384$ is the codebook size. The spatial downsampling factor of
eight yields $N=256$ discrete codes per modality. Each codebook index maps one-to-one to an atomic VLM token. Two additional
tokens, \texttt{\detokenize{<|future_joint_vq8_start|>}} and
\texttt{\detokenize{<|future_joint_vq8_end|>}}, delimit the future-state
payload. The visual vocabulary $\mathcal{V}_{\mathrm{vis}}$ therefore
contains $K+2=16{,}386$ tokens.

\paragraph{Spatially aligned joint serialization}
Let $q_i^m$ denote the VLM token corresponding to the $i$-th code in the
row-major traversal of $Q^m_{t+\Delta}$. We interleave the three modalities
at each spatial location in RGB--depth--mask order:
\begin{equation}
        y^{\mathrm{vis}} = [b_{\mathrm{start}},
        q^{\mathrm{rgb}}_1,
        q^{\mathrm{depth}}_1,
        q^{\mathrm{mask}}_1,\ldots, 
        q^{\mathrm{rgb}}_N,
        q^{\mathrm{depth}}_N,
        q^{\mathrm{mask}}_N,
        b_{\mathrm{end}}],
    \label{eq:joint_visual_sequence}
\end{equation}
where $b_{\mathrm{start}}$ and $b_{\mathrm{end}}$ denote the two boundary
tokens. The sequence contains $3N=768$ VQ tokens and two boundary tokens,
for a total of 770 tokens. This ordering places tokens from corresponding
spatial locations next to one another, providing local cross-modal context
during autoregressive generation.

The model generates the future-state sequence autoregressively through the shared LM output head. No modality-specific prediction heads or pixel-space reconstruction losses are introduced during VLM training. At inference time, the generated payload is mapped back to codebook indices and de-interleaved into three $16\times16$ grids. Each grid is reconstructed separately using the same frozen VQ decoder to produce the corresponding RGB image, depth map, or robot mask.

\subsection{Unified Vocabulary and Training Objective}
\label{sec:unified_objective}

We jointly train embodied understanding, action generation, and visual-foundation generation using the unified vocabulary in Eq.~\eqref{eq:unified_vocab} and the autoregressive objective in Eq.~\eqref{eq:unified_ar}. Task formats specify the requested outputs, while loss masks select the corresponding target tokens. 

The three objectives provide complementary supervision for a shared physical representation. Embodied understanding captures task semantics and spatial relations; action generation models end-effector motion; and visual-foundation generation models future scene states. Joint training brings these signals into the same backbone, allowing the model to learn from both descriptions of physical tasks and direct motion and state targets. This is the motivation for unifying the objectives beyond sharing a token-generation interface.

\section{Data}
\label{sec:data}

We organize the training data into two stages: physical-aware pre-training
and embodied supervised fine-tuning (SFT). Both stages provide supervision
for perception and understanding, action generation, and future-state
prediction. Physical-aware pre-training derives these three types of
supervision from large-scale human interaction videos. Embodied SFT reuses the
pre-training data construction pipeline to incorporate more diverse data
sources, including interactions in real-robot and simulated environments,
for further fine-tuning on high-quality data. This data design supports
learning across the physical loop by connecting environmental perception
and understanding, action-based interaction, and changes in physical state.
Table~\ref{tab:data_curriculum} reports the training data volume for each
supervision type at each stage.

\begin{table*}[t]
    \centering
    \footnotesize
    \setlength{\tabcolsep}{5pt}
    \renewcommand{\arraystretch}{1.08}
    \begin{tabular}{
        >{\raggedright\arraybackslash}p{0.14\textwidth}
        >{\raggedright\arraybackslash}p{0.14\textwidth}
        >{\raggedright\arraybackslash}p{0.30\textwidth}
        >{\raggedright\arraybackslash}p{0.14\textwidth}
        >{\centering\arraybackslash}p{0.13\textwidth}}
        \toprule
        Stage & Component & Supervision & Domain & Training samples \\
        \midrule
        \multirow[c]{4}{0.14\textwidth}{\centering Pre-training}
        & Perception
        & Captioning, VQA, spatial labels
        & Human videos
        & 24.3M \\
        & Human action
        & EEF trajectories
        & Human videos
        & 31.2M \\
        & Future state
        & RGB, depth, and human masks
        & Human videos
        & 26.8M \\
        & General
        & Instruction following
        & Multimodal data
        & 14.9M \\
        \midrule
        \multirow[c]{4}{0.14\textwidth}{\centering Embodied SFT}
        & Understanding
        & Language and spatial responses
        & Multimodal data
        & 6.61M \\
        & Action
        & EEF trajectories
        & Human/robot/simulation
        & 5.4M \\
        & Future state
        & RGB, depth, and robot masks
        & Robot/simulation
        & 1.2M \\
        & General
        & Instruction following
        & Multimodal data
        & 1M \\
        \bottomrule
    \end{tabular}
    \caption{Composition of the two-stage training data. Physical-aware
    pre-training data are constructed from human interaction videos; embodied
    SFT combines curated human, real-robot, and simulation data. Both stages
    also include general language and vision-language instruction data.}
    \label{tab:data_curriculum}
\end{table*}

\subsection{Physical-Aware Pre-training Data}

Our pre-training corpus is built from human interaction videos that expose how
people perceive a task, act in the environment, and change its physical state.
We aggregate Xperience-10M~\citep{Ropedia2026Xperience10m},
Egocentric-10K~\citep{BuildAI2025Egocentric10k}, Ego4D~\citep{Grauman2022Ego4dAround},
EgoVerse~\citep{Punamiya2026EgoverseAn}, EgoDex~\citep{Hoque2025EgodexLearning}, EgoLife~\citep{Yang2025EgolifeTowards}, and
Ego-Exo4D~\citep{Grauman2023EgoExo4d}, together with two in-house corpora.
\textbf{PhysBrain-Human}~\citep{Lin2026HumanAs} contains egocentric recordings, including sequences
with synchronized exocentric views.
\textbf{PhysBrain-Ego360}~\citep{DeepCyboTeamndEgo360A} provides
panoramic videos of the surrounding environment together with full-body poses,
hand motion, and task-level speech recorded during activity execution. These
sources cover daily activities, tool use, human--object interaction, and
long-horizon tasks across egocentric, exocentric, and panoramic viewpoints. We segment long recordings at task and action boundaries to obtain
task-coherent episodes containing a high-level goal and its execution process.
We discard low-quality episodes with severe motion blur, substantial
underexposure or overexposure, corrupted or missing frames, or key interactions
that are obscured or outside the field of view.
The curated source corpus contains approximately 30,000 hours of video.
From this corpus, we construct the three types of training data described below.

\subsubsection{Physical Perception Data}

Physical perception data consist of structured annotations, captions, and
question-answer pairs derived from human interaction videos. Specialized
open-source and in-house models generate structured
pseudo-labels from sampled frames and clips. Image-level supervision covers
object detection, segmentation, depth estimation, pointing, counting, and 3D
object detection. Video-level supervision covers temporal grounding,
spatio-temporal grounding, and object tracking. Synchronized egocentric,
exocentric, and panoramic recordings additionally support cross-view object
correspondence and referring grounding under changes in viewpoint, scale, and
visibility.

We complement these structured targets with physically grounded captioning and
question answering. Frame-level captions describe visible objects and
interaction-relevant scene properties; segment-level captions align
fine-grained actions with manipulated objects and observed state changes; and
episode-level captions summarize the environment, task goal, and execution
progress. For PhysBrain-Ego360, we combine synchronized speech transcripts with
video evidence to derive task descriptions and temporally grounded step-level
captions. Physical VQA examples probe task recognition, progress assessment,
action understanding, human--object interaction, temporal order, and observed
state changes. We retain the temporal or multi-view evidence required by each
target and reject pseudo-labels, captions, and answers that are inconsistent
with the source media or supporting annotations. The resulting physical
perception corpus contains 24.3M training samples.

\subsubsection{Human Action Data}

We construct human action data from motion trajectories aligned with task
execution in human interaction videos. Using the Human-as-Humanoid
pipeline~\citep{Lin2026HumanAs}, we recover human motion and derive wrist
end-effector trajectories from the recovered kinematic chain. Temporal
correspondences between the video and motion sequences associate these
trajectories with task descriptions and visual observations, providing task
and scene context for each motion segment.

We divide interaction videos into short clips and extract continuous motion
trajectories and their corresponding observations. Each sample preserves the
temporal relationships among an action segment, the subsequent visual
observation, and the continuation of the action, capturing a continuous
interaction process. We check pose-recovery reliability, video--trajectory
alignment, and motion plausibility, rejecting samples with unreliable
reconstruction, missing temporal correspondences, or implausible motion.
The resulting human action corpus contains 31.2M training samples.

\subsubsection{Human-Interaction Future-State Data}

We construct future-state data from human interaction videos by pairing the
scene context before an interaction with a subsequent physical state. We
divide videos into short clips and select a current observation and a target future
observation to form future-state prediction samples. For
interaction sequences associated with action trajectories, we also preserve
the temporal correspondences between action segments and subsequent
observations, linking state annotations to motion within the same interaction.

Each future-state target comprises a spatially aligned RGB image, depth map,
and human-part segmentation mask. The RGB image is taken directly from the
video frame at the target timestamp. The depth map and segmentation mask
covering the interacting hands and arms are generated by the corresponding
perception pipelines. These annotations capture scene appearance, depth
structure, and the human regions involved in the interaction, respectively,
and describe the physical state at a common timestamp.

We validate the temporal offsets of observation pairs, the integrity of images
and annotations, the numeric validity of depth maps and masks, and spatial
correspondences across modalities. Samples with temporal mismatches, missing
data, or inconsistent cross-modal alignment are discarded. The resulting
human-interaction future-state corpus contains 26.8M training samples.

\paragraph{General instruction data}
The general language and vision-language instruction data used for
pre-training comprise 14.9M samples.
Language sources include the FLAN Collection~\citep{Longpre2023TheFlan},
UltraChat~\citep{Ding2023EnhancingChat}, OpenAssistant Conversations~\citep{Kopf2023OpenassistantConversations},
selected subsets of UltraData-SFT-2605~\citep{OpenBMB2026UltradataSft},
MetaMathQA~\citep{Yu2023MetamathBootstrap}, and Magicoder-OSS-Instruct~\citep{Wei2024MagicoderEmpowering}.
Vision-language sources include FineVision~\citep{Wiedmann2025FinevisionOpen},
LLaVA-OneVision-1.5-Instruct-Data~\citep{An2025LlavaOnevision},
Cambrian-7M~\citep{Tong2024Cambrian1}, PixMo~\citep{Deitke2025MolmoAnd},
ShareGPT4Video~\citep{Chen2024Sharegpt4videoImproving}, and
LLaVA-Video-178K~\citep{Zhang2024VideoInstruction}. We perform cross-source
deduplication before capability-aware sampling because several collections
aggregate overlapping public data.

\subsection{Embodied Supervised Fine-Tuning Data}

The embodied SFT corpus combines high-quality human interaction data,
real-robot trajectories, and simulated interactions. From these sources, we
construct language and spatial annotations for embodied understanding,
end-effector trajectories for action generation, and aligned visual targets
for future-state prediction. To retain broad language and multimodal
capabilities during embodied adaptation, the mixture also includes 1M
high-quality general language and vision-language instruction samples.

\subsubsection{Embodied Understanding Data}

The embodied-understanding corpus combines the datasets listed in
Table~\ref{tab:embodied_understanding_data}, grouped by annotation type.
Across groups, native annotations are converted into
instruction-formatted language or spatial targets. We preserve temporal,
multi-view, and geometric evidence where required, standardize coordinates and
output formats, and filter examples for semantic validity, referential clarity,
geometric reliability, media integrity, and cross-source duplication.

\begin{table*}[t]
    \centering
    \scriptsize
    \setlength{\tabcolsep}{4pt}
    \renewcommand{\arraystretch}{1.05}
    \begin{tabular}{
        >{\centering\arraybackslash\bfseries}m{0.16\textwidth}
        >{\centering\arraybackslash}m{0.25\textwidth}
        >{\centering\arraybackslash}m{0.39\textwidth}
        >{\centering\arraybackslash}m{0.10\textwidth}}
        \toprule
        Capability group & Supervision & Representative sources & Scale \\
        \midrule
        Foundational visual--spatial perception
        & Counting, relative depth and distance, metric distance, spatial relations, and local 3D geometry
        & TallyQA~\cite{Acharya2019TallyqaAnswering}, DIW~\cite{Chen2016SingleImage}, VSR~\cite{Liu2023VisualSpatial},
          COCO~\cite{Lin2014MicrosoftCoco}, ADE20K~\cite{Zhou2017SceneParsing}, Omni3D~\cite{Brazil2023Omni3dA},
          Matterport3D~\cite{Chang2017Matterport3dLearning}, Argoverse~2~\cite{Wilson2021Argoverse2},
          and ScanNet~\cite{Dai2017ScannetRichly}
        & 500K \\
        \addlinespace[3pt]
        \rowcolor{amappaleblue!60}
        Spatial and multi-view understanding
        & Geometric relations, scene structure, viewpoint reasoning, cross-view correspondence, and 3D mental modeling
        & SSI-800K (SenseNova-SI)~\cite{Cai2026ScalingSpatial}, SpatialLadder~\cite{Li2026SpatialladderProgressive},
          SAT-Train~\cite{Ray2025SatDynamic}, SpatialMQA~\cite{Liu2025CanMultimodal},
          GRiD-3D~\cite{Lee2022WhatIs}, SpatialReasoner training data~\cite{Ma2025SpatialreasonerTowards},
          VSI-590K~\cite{Yang2026CambrianS}, VSI-Train-10K~\cite{Brown2025BenchmarkDesigners},
          SIMS-VSI-200K~\cite{Brown2025SimsV}, MindCube-Train~\cite{Wang2026MindcubeSpatial},
          and VST~\cite{Yang2025VisualSpatial}
        & 2.36M \\
        \addlinespace[3pt]
        Embodied reasoning and planning
        & Task state, progress, next action, procedural order, failure diagnosis, and corrective behavior
        & RoboVQA-Train~\cite{Sermanet2024RobovqaMultimodal}, HoloAssist~\cite{Wang2023HoloassistAn},
          EgoPlan-IT~\cite{Chen2023EgoplanBench}, RoboFail~\cite{Liu2023ReflectSummarizing},
          BridgeData~V2~\cite{Walke2023BridgedataV2}, AgiBot~\cite{AgiBotWorldContributors2025AgibotWorld},
          VLABench~\cite{Zhang2025VlabenchA}, and the human-interaction corpus
        & 1.65M \\
        \addlinespace[3pt]
        \rowcolor{amappaleblue!60}
        Grounding and affordance
        & Object and part localization, pointing, free-space and placement grounding, contact regions, and functional affordance
        & RoboRefIt-Train~\cite{Lu2023VlGrasp}, Objects365~\cite{Shao2019Objects365A},
          OpenImages~\cite{Kuznetsova2020TheOpen}, LVIS~\cite{Gupta2019LvisA},
          Visual Genome~\cite{Krishna2017VisualGenome}, CoSyn-Point~\cite{Yang2025ScalingText},
          PixMo-Points~\cite{Deitke2025MolmoAnd},
          Molmo2-MultiImagePoint~\cite{Clark2026Molmo2Open},
          RefSpatial-Train~\cite{Zhou2025RoboreferTowards},
          RoboPoint~\cite{Yuan2024RobopointA}, RoboSpatial-Train~\cite{Song2025RobospatialTeaching},
          RoboAfford-Train~\cite{Tang2025RoboaffordA}, HANDAL~\cite{Guo2023HandalA},
          PACO~\cite{Ramanathan2023PacoParts}, InstructPart~\cite{Wan2025InstructpartTask}, and
          ShareRobot~\cite{Ji2025RobobrainA}
        & 1.8M \\
        \addlinespace[3pt]
        Visual trajectory reasoning
        & Image-space traces and waypoints for approach, contact, transport, and placement
        & FSD~\cite{Yuan2026FromSeeing}, ShareRobot~\cite{Ji2025RobobrainA},
          Open X-Embodiment~\cite{OpenXEmbodimentCollaboration2023OpenX}, DROID~\cite{Khazatsky2024DroidA},
          ManiSkill~\cite{Mu2021ManiskillGeneralizable}, RoboCasa~\cite{Nasiriany2024RobocasaLarge},
          Molmo2-VideoPoint~\cite{Clark2026Molmo2Open},
          MolmoPoint-TrackAny/TrackSyn~\cite{Clark2026MolmopointBetter},
          Ego4D~\cite{Grauman2022Ego4dAround}, Egocentric-10K~\cite{BuildAI2025Egocentric10k},
          EgoDex~\cite{Hoque2025EgodexLearning}, PhysBrain-Ego360~\cite{DeepCyboTeamndEgo360A}, and
          PhysBrain-Human~\cite{Lin2026HumanAs}
        & 300K \\
        \bottomrule
    \end{tabular}
    \caption{Composition of the embodied-understanding data. Scales report
    the number of training samples in each capability group.}
    \label{tab:embodied_understanding_data}
\end{table*}

Foundational perception data are organized as instruction-following samples
by converting native annotations into task instructions paired with textual
or spatial targets. These samples cover counting, depth and distance
judgments, spatial relation understanding, and object localization. Depending
on the task, they take the form of question answering, captioning, or
structured spatial prediction, with object categories, numerical ranges, and
answer distributions balanced during construction.
Spatial and multi-view examples retain their original image,
video, camera, and scene organization so that viewpoint-dependent evidence is
not collapsed during conversion. Reasoning and planning data combine original
clips with temporally ordered frame sequences.
Grounding and visual-trajectory data use a shared normalized spatial format for
points, boxes, regions, and image-space waypoints.

\subsubsection{Embodied Action Data}
\label{sec:embodied_action_data}

We construct the embodied-action corpus by combining trajectories from 17
real-robot and simulation sources with high-quality human motion trajectories.
These data span diverse embodiments, viewpoints, tasks, and motion
distributions. As summarized in Table~\ref{tab:action_data_scale}, the raw
collection contains approximately 2,620 hours of real-robot and simulation
trajectories and 500 hours of high-quality human motion data.

The source datasets differ in action semantics, physical units, and temporal
sampling conventions. We first determine whether each source records observed
states or commanded targets and extract motion sequences according to these
semantics. We standardize physical units and gripper open--close conventions
while retaining each source's audited native coordinate axes. For human data,
we reuse the Human-as-Humanoid pipeline~\citep{Lin2026HumanAs} employed during
pre-training to recover wrist trajectories, retaining sequences with reliable
pose recovery and temporal alignment. We then synchronize trajectories with
visual observations and task descriptions to provide task and scene context
for each motion sequence.

\begin{table}[t]
    \centering
    \small
    \setlength{\tabcolsep}{6pt}
    \renewcommand{\arraystretch}{1.05}
    \begin{tabular}{
        >{\centering\arraybackslash\bfseries}m{0.23\linewidth}
        >{\centering\arraybackslash}m{0.55\linewidth}
        >{\centering\arraybackslash}m{0.12\linewidth}}
        \toprule
        Source family & Datasets & Scale \\
        \midrule
        Bimanual platforms
        & OpenGalaxea~\cite{Jiang2025GalaxeaOpen}, AgiBotWorld~\cite{AgiBotWorldContributors2025AgibotWorld},
          GM100/CobotMagic~\cite{RobbyantndGm100A},
          RoboTwin~\cite{Mu2025RobotwinDual}, and GR00T-X~\cite{NVIDIAndPhysicalaiRobotics}
        & 987 h \\
        \addlinespace[3pt]
        \rowcolor{amappaleblue!60}
        Single-arm platforms
        & DROID~\cite{Khazatsky2024DroidA}, FurnitureBench~\cite{Heo2023FurniturebenchReproducible},
          FMB~\cite{Luo2024FmbA}, Fractal/RT-1~\cite{Brohan2022Rt1},
          BC-Z~\cite{Jang2022BcZ}, BridgeData~V2~\cite{Walke2023BridgedataV2},
          Dobb-E~\cite{Shafiullah2023OnBringing},
          TACO Play~\cite{RoseteBeas2023LatentPlans,Mees2023GroundingLanguage}, and
          Berkeley Cable Routing~\cite{Luo2023MultiStage}
        & 1,339 h \\
        \addlinespace[3pt]
        Simulation benchmarks
        & RoboCasa~\cite{Nasiriany2024RobocasaLarge}, LIBERO~\cite{Liu2023LiberoBenchmarking},
          and VLA-Arena~\cite{Zhang2025VlaArena}
        & 294 h \\
        \addlinespace[3pt]
        \rowcolor{amappaleblue!60}
        Human interaction
        & Selected high-quality human motion trajectories
        & 500 h \\
        \bottomrule
    \end{tabular}
    \caption{Scale and composition of the embodied-action data used for
    supervised fine-tuning. The corpus combines real-robot and simulation
    trajectories with 500 hours of high-quality human motion data.}
    \label{tab:action_data_scale}
\end{table}

For selected sources with low sampling rates, we resample continuous
trajectories to improve temporal resolution. Resampling preserves the original
sample anchors without introducing additional candidate windows, maintaining
the relative sampling weights of the source datasets. We reject samples with
invalid poses, rotations, timestamps, media, or annotations, and further filter
trajectories using the motion-informativeness criterion from
FrameSkip~\citep{Yu2026FrameskipLearning}. Uninformative segments are downsampled, while
segments containing new motion or interaction changes are retained. Stationary
segments are selectively retained to represent meaningful pauses during task
execution. Each final sample includes the task description, visual
observations, embodiment metadata, action frequency, and temporally aligned
motion trajectories. All data splits are constructed at the episode level.

\subsubsection{Embodied Future-State Data}
\label{sec:visual_foundation_data}

We construct embodied future-state data from the real-robot and simulation
trajectories used in the action corpus, associating task descriptions and
current visual observations with subsequent target frames. To accommodate
differences in sampling rates and recording conventions, we use timestamps
and each source's native frame rate to establish temporal correspondences.
Target frames are drawn from the same continuous interaction sequence and
aligned with the corresponding task context.

We follow the multimodal state-annotation procedure used for human-interaction
pre-training to construct spatially aligned RGB images, depth maps, and
segmentation masks for each target frame. RGB images are taken directly from
the source trajectory, metric z-depth maps are generated with
MoGe-2~\citep{Wang2025Moge2}, and robot-body masks are obtained with
RoboEngine~\citep{Yuan2025RoboenginePlug}. The segmentation targets cover the robot
body, extending the human hand-and-arm annotations used during pre-training
to robot embodiments. These annotations capture scene appearance, spatial
depth, and the regions occupied by the robot, respectively, and describe a
physical state at a common timestamp in the same image coordinate system.

We check the completeness of target frames and annotations, validate the
temporal offsets of observation pairs and the numeric validity of depth maps
and masks, and verify spatial correspondences across modalities. Samples with
missing data, temporal mismatches, or inconsistent annotation alignment are
discarded, ensuring that the retained RGB images, depth maps, and robot masks
form consistent future-state targets. The resulting embodied future-state
corpus contains 1.2M training samples.

\section{Training}
\label{sec:training}

We train PhysBrain 1.5 using ms-swift\footnote{\url{https://github.com/modelscope/ms-swift}} with its Megatron-Core~\citep{Shoeybi2019MegatronLm} backend, taking Qwen3-VL-Instruct (8B) as the base model and extending its vocabulary with action and visual-state tokens as described in Sec.~\ref{sec:model}. Training proceeds in two stages: physical-aware pre-training followed by supervised fine-tuning, with the second stage initialized from the first-stage checkpoint. We train for one epoch in each stage using the same hyperparameter configuration, mixing data categories in proportion to their sample counts reported in Sec.~\ref{sec:data}. Both stages jointly optimize text generation, action-token generation, and future visual-state generation under the unified autoregressive objective.

For both stages, we use the distributed Adam optimizer with
$\beta_1=0.9$, $\beta_2=0.95$, $\epsilon=10^{-8}$,
and weight decay of $0.1$.
In each stage, the learning rate warms up over the first
$3\%$ of training steps to a peak of $2\times10^{-5}$,
followed by cosine decay.
The maximum sequence length is 32,768 tokens,
with sequence packing enabled to improve token utilization.
Each global batch contains approximately 2K training samples
on average after packing.
We clip the gradient norm at $1.0$ and train in bfloat16 precision.

\section{Evaluation}
\label{sec:eval}

\begingroup
\protected\def\pbTableBold#1{{\rmfamily\bfseries #1}}
\let\textbf\pbTableBold
\definecolor{pbBestBlue}{RGB}{172,207,239}
\definecolor{pbSecondBlue}{RGB}{211,230,248}
\newcommand{\pbClosed}[1]{\textcolor{gray}{#1}}
\newcommand{\pbBest}[1]{\cellcolor{pbBestBlue}\textbf{#1}}
\newcommand{\pbSecond}[1]{\cellcolor{pbSecondBlue}#1}
\newlength{\pbModelHeaderWidth}
\newlength{\pbModelHeaderHalfHeight}
\newcommand{\pbModelHeader}[1]{\raisebox{-0.5\height}[\pbModelHeaderHalfHeight][\pbModelHeaderHalfHeight]{\rotatebox{90}{\makebox[\pbModelHeaderWidth][c]{\makecell[c]{#1}}}}}
\begin{table}[!t]
\centering
\setlength{\belowcaptionskip}{4pt}
\caption{\textbf{Results on Embodied Understanding.}
All results are reported on a 0--100 scale, with higher values indicating better performance.
Closed-source models are shown in gray and excluded from ranking.
Darker blue cells with bold text and lighter blue cells mark the best and second-best open-source results, respectively.}
\label{tab:embodied_benchmarks}
\setlength{\tabcolsep}{2.15pt}
\renewcommand{\arraystretch}{1.04}
\scriptsize
\settowidth{\pbModelHeaderWidth}{Hy-Emb.-VLM-1.0}
\setlength{\pbModelHeaderHalfHeight}{0.5\pbModelHeaderWidth}
\addtolength{\pbModelHeaderHalfHeight}{2pt}
\begin{adjustbox}{max width=\textwidth}
\begin{tabular}{@{}>{\raggedright\arraybackslash}m{2.85cm}>{\raggedright\arraybackslash}m{2.35cm}|ccc|ccccccccc@{}}
\toprule
\multicolumn{2}{c|}{} &
\multicolumn{3}{c|}{\textcolor{gray}{\makecell{Closed-Source\\Models}}} &
\multicolumn{9}{c}{\makecell{Open-Source\\Models}} \\
\cmidrule(lr){3-5}\cmidrule(lr){6-14}
\multicolumn{1}{@{}l}{\raisebox{-0.5\height}{\textbf{Category}}} &
\multicolumn{1}{l|}{\raisebox{-0.5\height}{\textbf{Benchmarks}}} &
\textcolor{gray}{\pbModelHeader{\strut Gemini 3.6 Flash\\\strut minimal think.}} &
\textcolor{gray}{\pbModelHeader{\strut GPT 6 Astra\\\strut low think.}} &
\textcolor{gray}{\pbModelHeader{\strut Claude Opus 5\\\strut adapt. low think.}} &
\pbModelHeader{\strut Hy-Emb.-VLM-1.0\\\strut 30A3B think.} &
\pbModelHeader{\strut Embodied-R1.5\\\strut 8B} &
\pbModelHeader{\strut RynnBrain1.1\\\strut 9B} &
\pbModelHeader{\strut RoboBrain2.5\\\strut 8B} &
\pbModelHeader{\strut MiMo Embodied\\\strut 7B think.} &
\pbModelHeader{\strut Cosmos3 Nano\\\strut 8B+8B} &
\pbModelHeader{\strut ACE-Brain-0.5\\\strut 8B} &
\pbModelHeader{\strut Qwen3-VL-Inst.\\\strut 8B} &
\pbModelHeader{\strut \textbf{PhysBrain 1.5}\\\strut \textbf{8B}} \\
\midrule
\multirow{2}{*}{\makecell[l]{Foundational\\Visual-Spatial Percept.}} & BLINK & \pbClosed{89.2} & \pbClosed{80.6} & \pbClosed{85.5} & \pbSecond{86.1} & 78.6 & 84.8 & 80.4 & 69.5 & 80.6 & 77.8 & 81.7 & \pbBest{87.9} \\
 & CV-Bench & \pbClosed{90.0} & \pbClosed{84.9} & \pbClosed{87.8} & 89.3 & 86.9 & 88.1 & 88.0 & \pbSecond{89.4} & 88.3 & 85.0 & 85.8 & \pbBest{90.0} \\
\midrule
\multirow{9}{*}{\makecell[l]{Spatial and\\Multi-view\\Understanding}} & 3DSRBench & \pbClosed{67.7} & \pbClosed{62.3} & \pbClosed{63.3} & \pbSecond{62.1} & 52.3 & 59.0 & 54.7 & 53.2 & 53.5 & 51.8 & 53.6 & \pbBest{63.0} \\
 & EmbSpatial-Bench & \pbClosed{84.7} & \pbClosed{79.8} & \pbClosed{81.1} & 80.5 & 75.2 & \pbSecond{81.9} & 77.9 & 77.0 & \pbBest{82.1} & 77.9 & 79.9 & 81.8 \\
 & MindCube & \pbClosed{77.1} & \pbClosed{78.8} & \pbClosed{66.7} & 65.4 & 30.9 & 82.1 & 30.4 & 31.3 & 39.2 & \pbBest{94.2} & 33.9 & \pbSecond{86.2} \\
 & MMSI-Bench & \pbClosed{52.8} & \pbClosed{57.9} & \pbClosed{44.1} & 39.0 & 30.1 & \pbBest{46.0} & 29.4 & 31.8 & 35.7 & 35.8 & 30.8 & \pbSecond{41.0} \\
 & Q-Spatial-Bench & \pbClosed{87.1} & \pbClosed{68.3} & \pbClosed{74.3} & \pbSecond{76.2} & 51.5 & 66.3 & \pbSecond{76.2} & 47.5 & 46.5 & 35.6 & 54.5 & \pbBest{81.2} \\
 & RoboSpatial-Home & \pbClosed{71.0} & \pbClosed{73.7} & \pbClosed{72.0} & 71.2 & \pbSecond{72.0} & 70.4 & 67.3 & 60.1 & 66.4 & 65.9 & 67.7 & \pbBest{73.9} \\
 & SAT & \pbClosed{88.7} & \pbClosed{96.7} & \pbClosed{88.7} & \pbBest{81.3} & 74.0 & 78.0 & 65.3 & 76.0 & 69.3 & \pbSecond{79.3} & 70.7 & \pbSecond{79.3} \\
 & VSI-Bench & \pbClosed{51.8} & \pbClosed{59.8} & \pbClosed{21.3} & 55.7 & 56.3 & \pbBest{65.9} & 43.9 & 45.0 & 50.4 & 57.9 & 55.1 & \pbSecond{61.9} \\
 & ViewSpatial-Bench & \pbClosed{56.6} & \pbClosed{54.2} & \pbClosed{49.8} & 52.4 & 43.9 & \pbSecond{56.4} & 39.3 & 39.7 & 55.0 & 48.3 & 40.3 & \pbBest{62.5} \\
\midrule
\multirow{6}{*}{\makecell[l]{Embodied\\Cognition,\\Reasoning, and\\Planning}} & COSMOS & \pbClosed{65.3} & \pbClosed{75.7} & \pbClosed{71.0} & 64.5 & \pbSecond{67.5} & 64.3 & 57.7 & 56.5 & 66.8 & 46.5 & 60.6 & \pbBest{72.8} \\
 & EgoPlan-Bench2 & \pbClosed{53.1} & \pbClosed{69.3} & \pbClosed{48.4} & 47.5 & \pbSecond{53.1} & 37.0 & 33.8 & 37.9 & 42.6 & 33.5 & 32.1 & \pbBest{62.1} \\
 & ERQA & \pbClosed{72.3} & \pbClosed{75.8} & \pbClosed{61.5} & \pbBest{56.3} & 42.3 & 46.5 & 44.3 & 44.8 & 47.5 & 45.3 & 42.5 & \pbSecond{52.8} \\
 & ERQA-PLUS & \pbClosed{92.2} & \pbClosed{86.1} & \pbClosed{88.3} & 82.2 & 80.7 & \pbSecond{83.6} & 81.1 & 81.1 & 83.5 & 81.4 & 82.5 & \pbBest{85.2} \\
 & RoboVQA & \pbClosed{41.7} & \pbClosed{37.0} & \pbClosed{36.0} & 41.2 & 60.8 & \pbSecond{60.9} & 48.4 & 55.5 & 51.5 & 44.7 & 58.5 & \pbBest{61.5} \\
 & VLABench & \pbClosed{64.7} & \pbClosed{66.3} & \pbClosed{60.3} & \pbSecond{50.9} & 40.9 & 42.4 & 37.6 & 39.9 & 48.7 & 29.1 & 44.8 & \pbBest{76.4} \\
\midrule
\multirow{9}{*}{\makecell[l]{Spatial\\Grounding,\\Pointing, and\\Affordance}} & Part-Affordance & \pbClosed{64.7} & \pbClosed{55.0} & \pbClosed{78.1} & 62.8 & \pbSecond{83.4} & 40.5 & 24.7 & 61.9 & 33.3 & 36.3 & 56.4 & \pbBest{84.0} \\
 & PIOBench & \pbClosed{80.9} & \pbClosed{79.7} & \pbClosed{71.6} & 62.6 & 61.5 & \pbSecond{64.8} & 62.3 & 56.2 & 64.3 & 60.4 & 53.7 & \pbBest{68.3} \\
 & PixMo-Points & \pbClosed{74.8} & \pbClosed{75.2} & \pbClosed{56.7} & 55.8 & \pbSecond{65.3} & 29.9 & 57.3 & 51.5 & 60.5 & \pbBest{65.6} & 53.7 & 62.2 \\
 & PointBench & \pbClosed{69.6} & \pbClosed{71.9} & \pbClosed{68.6} & 61.1 & \pbBest{64.5} & 45.5 & 61.7 & 51.6 & 62.8 & 63.0 & 61.5 & \pbSecond{64.3} \\
 & RefSpatial-Bench & \pbClosed{76.4} & \pbClosed{78.0} & \pbClosed{56.3} & 47.4 & 55.2 & \pbBest{63.2} & \pbSecond{59.9} & 40.0 & 52.8 & 53.3 & 43.6 & 50.9 \\
 & RoboAfford & \pbClosed{83.4} & \pbClosed{84.6} & \pbClosed{82.3} & 75.8 & 76.9 & 74.0 & 71.7 & 71.8 & \pbBest{81.1} & 71.8 & 66.8 & \pbSecond{80.4} \\
 & RoboRefit & \pbClosed{85.9} & \pbClosed{85.1} & \pbClosed{83.5} & 82.5 & 86.1 & 81.9 & 82.8 & 75.5 & \pbSecond{86.2} & 84.6 & 81.3 & \pbBest{89.6} \\
 & VABench-Point & \pbClosed{60.7} & \pbClosed{65.3} & \pbClosed{64.7} & 60.5 & \pbBest{75.7} & 15.8 & 10.6 & 47.7 & 53.2 & 5.3 & 46.3 & \pbSecond{65.2} \\
 & Where2Place & \pbClosed{73.4} & \pbClosed{76.9} & \pbClosed{69.0} & 65.4 & \pbBest{75.0} & 72.7 & 72.0 & 63.9 & \pbSecond{74.4} & 56.0 & 64.7 & 72.1 \\
\midrule
\multirow{2}{*}{\makecell[l]{Visual Trace and\\Traj. Reasoning}} & ShareRobot-Traj. & \pbClosed{82.8} & \pbClosed{83.1} & \pbClosed{81.9} & 84.5 & 84.0 & 79.3 & \pbBest{85.2} & 69.0 & 80.9 & 82.1 & 78.2 & \pbSecond{84.9} \\
 & VABench-V.-Trace & \pbClosed{84.2} & \pbClosed{91.6} & \pbClosed{88.9} & 87.5 & \pbBest{92.3} & 86.6 & 85.4 & 82.1 & 88.3 & 82.7 & 84.4 & \pbSecond{89.8} \\
\midrule
\multicolumn{2}{l|}{\textbf{Overall Average}} & \pbClosed{73.0} & \pbClosed{73.3} & \pbClosed{67.9} & \pbSecond{66.0} & 64.9 & 63.1 & 58.2 & 57.4 & 62.3 & 59.0 & 59.5 & \pbBest{72.5} \\
\bottomrule
\end{tabular}
\end{adjustbox}
\end{table}
\endgroup

Embodied behavior can be viewed as a recurrent physical loop: the agent
perceives a 2D scene, constructs a 3D spatial representation, plans an action
strategy, localizes the target point, generates a motion trajectory, executes
the action, and observes the resulting state transition. Guided by this view,
we evaluate PhysBrain 1.5 from four complementary perspectives. First, the
capabilities underlying the first five stages are assessed on 28 embodied
benchmarks organized into five groups: visual-spatial perception, spatial and
multi-view understanding, embodied cognition and planning, spatial grounding and affordance, and visual trace and trajectory reasoning. Second, we evaluate general multimodal understanding to assess the model’s broad perception, reasoning, and commonsense capabilities, which complement its embodied skills and provide a foundation for compositional generalization across diverse physical scenarios. We further show the model's ability to generate
action-trajectory tokens and to predict future visual states, corresponding to
action execution and state transition in the physical loop. Together, these
evaluations trace the path from multimodal perception and embodied
understanding to planning, action generation, and prediction of the evolving
physical world.

\begin{figure}
    \centering
    \includegraphics[width=1.0\linewidth]{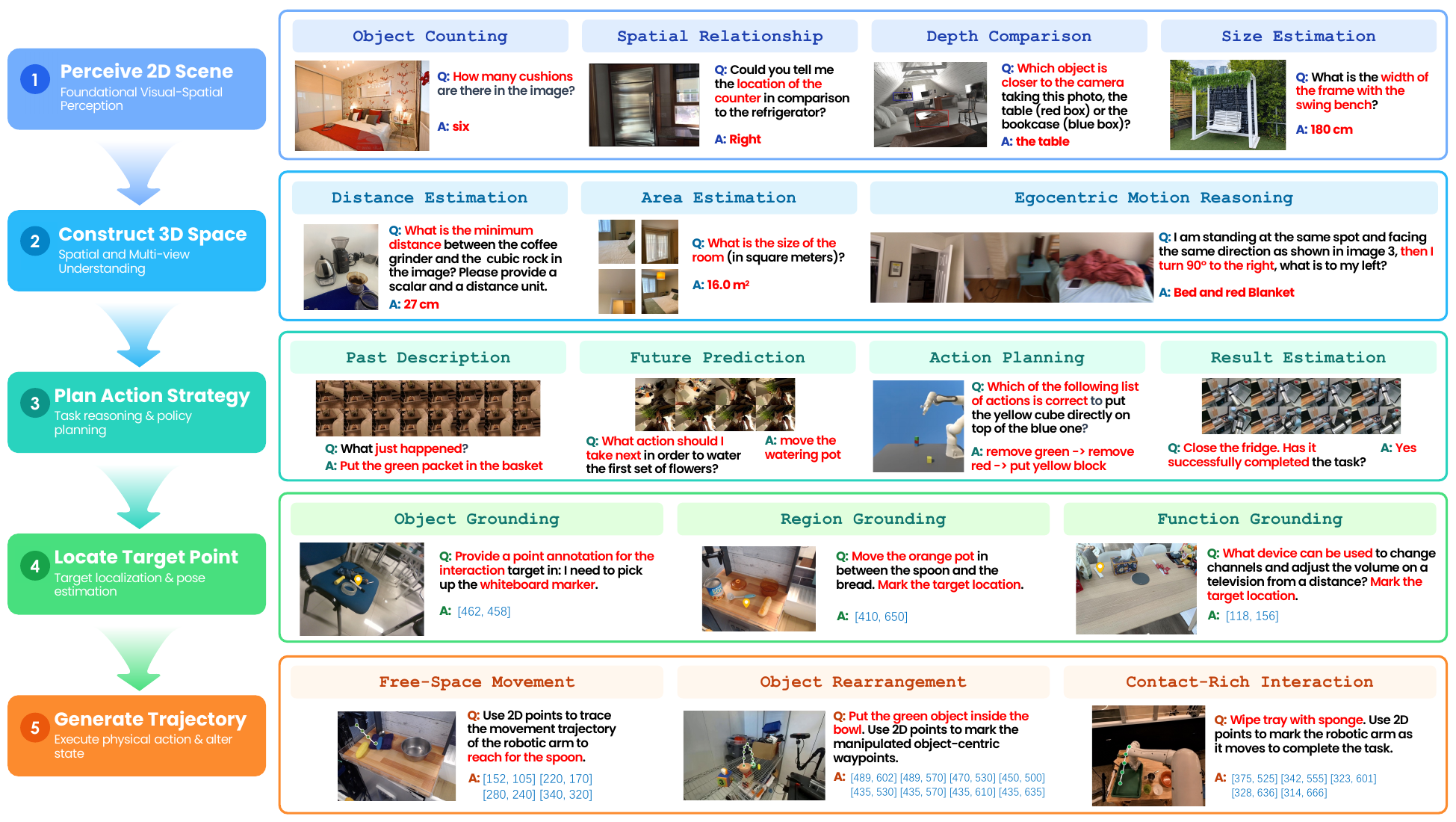}
    \caption{\textbf{Qualitative examples of embodied understanding with PhysBrain 1.5.} Representative cases illustrate visual-spatial perception, 3D and multi-view understanding, embodied reasoning and planning, spatial grounding and affordance, and visual trace and trajectory reasoning. Responses include language answers, spatial coordinates, and waypoint sequences, with predicted points and trajectories overlaid on the corresponding observations.}
    \label{fig:evaluation-cases}
\end{figure}

\subsection{Embodied Understanding}
\subsubsection{Benchmarks}
We evaluate embodied understanding on 28 benchmarks spanning five complementary capability groups. Foundational Visual-Spatial Perception assesses basic visual recognition and spatial discrimination using BLINK~\cite{Fu2024BlinkMultimodal} and CV-Bench~\cite{Tong2024Cambrian1}. Spatial and Multi-view Understanding evaluates geometry, spatial relations, viewpoint changes, and cross-view reasoning using 3DSRBench~\cite{Ma20253dsrbenchA}, EmbSpatial-Bench~\cite{Du2024EmbspatialBench}, MindCube~\cite{Wang2026MindcubeSpatial}, MMSI-Bench~\cite{Yang2026MmsiBench}, Q-Spatial-Bench~\cite{Liao2024ReasoningPaths}, RoboSpatial-Home~\cite{Song2025RobospatialTeaching}, SAT~\cite{Ray2025SatDynamic}, VSI-Bench~\cite{Yang2025ThinkingIn}, and ViewSpatial-Bench~\cite{Li2025ViewspatialBench}. Embodied Cognition, Reasoning and Planning measures goal-directed reasoning, procedural understanding, and planning using COSMOS~\cite{NVIDIA2025CosmosReason1}, EgoPlan-Bench2~\cite{Qiu2026EgoplanBench2}, ERQA~\cite{GeminiRoboticsTeam2025GeminiRobotics}, ERQA-PLUS~\cite{Yang2026ErqaPlus}, RoboVQA~\cite{Sermanet2024RobovqaMultimodal}, and VLABench~\cite{Zhang2025VlabenchA}. Spatial Grounding, Pointing, and Affordance evaluates object localization, referring, affordance understanding, and placement reasoning using Part-Affordance-2K~\cite{Yuan2025EmbodiedR1}, PIOBench S1 \& S2~\cite{Xue2025PointIt}, PixMo-Points~\cite{Deitke2025MolmoAnd}, PointBench~\cite{Cheng2025PointarenaProbing}, RefSpatial-Bench~\cite{Zhou2025RoboreferTowards}, RoboAfford~\cite{Hao2025RoboaffordA}, RoboRefit~\cite{Lu2023VlGrasp}, VABench-Point~\cite{Yuan2026FromSeeing}, and Where2Place~\cite{Yuan2024RobopointA}. Finally, Visual Trace and Trajectory Reasoning examines the understanding and prediction of motion traces and interaction trajectories using ShareRobot-Trajectory~\cite{Ji2025RobobrainA} and VABench-Visual-Trace~\cite{Yuan2026FromSeeing}. More details on benchmark-specific evaluation protocols and implementation settings are provided in the Appendix~\ref{app:embodied-eval-protocols}. This benchmark suite provides broad coverage of the perceptual, spatial, grounding, reasoning, and planning capabilities required for embodied decision-making.

\subsubsection{Comparison Models}
We compare PhysBrain 1.5 against representative proprietary models, including Gemini 3.6 Flash~\cite{GoogleDeepMind2026Gemini3}, GPT 6 Astra~\cite{OpenAI2026Gpt6}, and Claude Opus 5~\cite{Anthropic2026IntroducingClaude}, and strong open-source embodied MLLMs, including Hy-Embodied-VLM-1.0~\cite{Wang2026HyEmbodied}, Embodied-R1.5~\cite{Yuan2026EmbodiedR1}, RynnBrain1.1~\cite{Li2026Rynnbrain1}, RoboBrain2.5~\cite{Tan2026Robobrain2}, MiMo Embodied~\cite{Hao2025MimoEmbodied}, Cosmos3 Nano~\cite{NVIDIA2026Cosmos3}, and ACE-Brain-0.5~\cite{ACEBrainTeam2026AceBrain}. We additionally include Qwen3-VL-Instruct (8B)~\cite{Bai2025Qwen3Vl} as a general-purpose VLM baseline. Since different works may use inconsistent evaluation metrics for the same benchmark, their reported scores are not always directly comparable. We therefore independently re-evaluate all comparison models using a single canonical metric for each benchmark; accordingly, some scores in our table may differ from those reported in the original papers. When a model’s official release specifies a benchmark-specific input format or recommended prompt, we follow it; otherwise, we use a default task prompt and input template. For consistency across models, we standardize the source-image resolution and the number of sampled video frames supplied to each model while retaining its native preprocessing pipeline, including any internal resizing or downsampling. Since most models in our comparison do not use an explicit thinking mode, we evaluate the proprietary models using the lowest available official thinking setting: minimal thinking for Gemini, low thinking for GPT, and adaptive thinking with low effort for Claude. Following official recommendations, we enable thinking for Hy-Embodied-VLM-1.0 and MiMo Embodied, while evaluating all other models without a thinking mode. For point-localization and grounding tasks, we adopt each model’s officially recommended point-output format and corresponding parsing procedure.

\subsubsection{Results and Analysis}

PhysBrain 1.5 demonstrates strong and well-rounded embodied understanding across the first five stages of our physical loop. As shown in Table~\ref{tab:embodied_benchmarks}, it achieves the highest overall average among the evaluated embodied open-source models, scoring 72.5 and surpassing the strongest open-source baseline, Hy-Embodied-VLM-1.0, by 6.5 points. It ranks first on 14 benchmarks and among the top two on 24. It also outperforms our base model, Qwen3-VL-Instruct (8B), on all 28 benchmarks. When benchmark scores are averaged with equal weight within each capability category, PhysBrain 1.5 ranks among the top two open-source models in all five categories, demonstrating broad strengths in perception, spatial reasoning, planning, grounding, and trajectory understanding. Despite its compact 8B scale, PhysBrain 1.5 approaches world-leading proprietary models such as Gemini 3.6 Flash (73.0) and GPT-6-Astra (73.3) in overall score, while outperforming Claude Opus 5 (67.9) by 4.6 points, highlighting its competitiveness in embodied understanding.

Figure~\ref{fig:evaluation-cases} illustrates these capabilities through representative examples organized along the physical loop. At the scene-perception stage, PhysBrain 1.5 counts objects, reasons about spatial relationships, compares relative depth, and estimates physical dimensions. Moving toward 3D spatial understanding, it estimates metric distances and room area and reasons about egocentric spatial relations under viewpoint changes. For action planning, it interprets past actions, predicts the next step, decomposes goals into ordered actions, and assesses task completion. Task intent is then grounded in actionable locations through object pointing, placement-region localization, and functional affordance identification. Finally, the model generates image-space waypoints for free-space movement, object rearrangement, and contact-rich interactions such as wiping. Together, these examples illustrate complementary capabilities for progressing from scene understanding to goal-directed interaction. More examples can be found in Appendix~\ref{app:qualitative_results}

\begin{table*}[t]
\centering
\small
\caption{\textbf{General multimodal understanding results (I).} Results cover broad image and video understanding and robustness to object hallucination. VideoMME uses up to 128 sampled frames without additional subtitles. MME is reported as the sum of the MME-P and MME-C sub-benchmark scores on its native point scale; all other scores are percentages. Higher values indicate better performance.}
\label{tab:general_multimodal_a}
\setlength{\tabcolsep}{6pt}
\begin{tabular}{cccccccc}
\toprule
\multicolumn{1}{c}{Model} & Size & MME & MMStar & RealWorldQA & VideoMME & MVBench & POPE (F1) \\
\midrule
Qwen3-VL-Inst. & 8B & 2392.70 & 64.99 & 68.37 & 69.07 & 68.53 & 88.40 \\
PhysBrain~1.5 & 8B & 2330.07 & 65.60 & 69.41 & 68.22 & 66.07 & 89.49 \\
\bottomrule
\end{tabular}

\par\vspace{1em}

\caption{\textbf{General multimodal understanding results (II).} Results cover diagram and chart understanding, document and scene-text reading, fine-grained visual recognition, and GUI grounding. All scores are percentages, with higher values indicating better performance.}
\label{tab:general_multimodal_b}
\setlength{\tabcolsep}{6pt}
\begin{tabular}{cccccccc}
\toprule
\multicolumn{1}{c}{Model} & Size & AI2D & ChartQA & DocVQA & TextVQA & V* & ScreenSpot \\
\midrule
Qwen3-VL-Inst. & 8B & 83.74 & 84.96 & 95.66 & 81.94 & 83.77 & 91.59 \\
PhysBrain~1.5 & 8B & 82.16 & 86.24 & 94.43 & 81.58 & 87.96 & 90.17 \\
\bottomrule
\end{tabular}
\end{table*}

\subsection{General Multimodal Understanding}
\label{sec:general_multimodal}

A capable embodied model should combine specialized physical skills with
broad multimodal understanding and commonsense reasoning. These general
capabilities provide context for interpreting observations, understanding
instructions, and relating events to task goals across diverse scenarios.
We therefore complement the embodied evaluations with a suite of general
image and video benchmarks, using Qwen3-VL-Instruct (8B)~\cite{Bai2025Qwen3Vl}
as a reference for general-purpose multimodal performance.

\subsubsection{Evaluation Protocol}
We evaluate 12 established and widely used benchmarks for general multimodal
understanding: MME~\citep{Fu2025MmeA}, MMStar~\citep{Chen2024AreWe}, RealWorldQA~\citep{xAI2024Grok1}, VideoMME~\citep{Fu2025VideoMme}, MVBench~\citep{Li2024MvbenchA}, POPE~\citep{Li2023EvaluatingObject}, AI2D~\citep{Kembhavi2016ADiagram}, ChartQA~\citep{Masry2022ChartqaA}, DocVQA~\citep{Mathew2021DocvqaA}, TextVQA~\citep{Singh2019TowardsVqa}, V*~\citep{Wu2024VGuided}, and ScreenSpot~\citep{Cheng2024SeeclickHarnessing}. This suite covers broad image
and video understanding, complemented by evaluations of object hallucination,
diagram and chart reasoning, document and scene-text understanding,
fine-grained visual recognition, and GUI grounding. Descriptions of the
individual benchmark scopes and task formats are provided in
Appendix~\ref{app:general-mm-protocol}.

We use lmms-eval~\cite{Zhang2024LmmsEval} to evaluate PhysBrain~1.5 and the original
Qwen3-VL-Instruct. Our primary comparison uses the locally re-evaluated
Qwen baseline, with the same task-specific prompts, visual preprocessing,
decoding settings, and scoring procedures for both models. For benchmarks
reported by Qwen3-VL, we align the evaluation with its publicly released
recipes as closely as the local implementation permits.\footnote{\url{https://github.com/QwenLM/Qwen3-VL/tree/main/evaluation}}
For benchmarks without a reported Qwen result, we use the default lmms-eval
task configuration. For both models, video inputs contain up to 128
sampled frames on VideoMME and 32 on MVBench. Complete decoding, visual preprocessing, prompt, and
answer-extraction settings are provided in Appendix~\ref{app:general-mm-protocol}.

\subsubsection{Results and Analysis}
PhysBrain~1.5 combines strong embodied performance with broad multimodal
understanding. As shown in Table~\ref{tab:general_multimodal_a}
and Table~\ref{tab:general_multimodal_b}, its overall performance across the
general evaluation suite is comparable to that of the base model,
Qwen3-VL-Instruct (8B). Its capabilities extend beyond specialized embodied
benchmarks to image and video understanding, visual reasoning, and the
interpretation of text, charts, and diagrams.

\subsection{Beyond Understanding: Action Modeling and Physical-World Prediction}

\subsubsection{Action Trajectory Token Prediction}

We examine action trajectory predictions from the post-trained model. Given the current RGB observation, a task instruction, and the preceding 16-step ground-truth action chunk, PhysBrain~1.5 autoregressively generates action tokens for the next 16 steps, which are decoded by ActionPiece into end-effector trajectories. Figure~\ref{fig:action_prediction_2d} visualizes the predicted and ground-truth positions in the image plane on test episodes unseen during training. Across the illustrated manipulation tasks, the predictions broadly follow the direction and shape of the reference motion. Examples involving object placement, tool use, and bimanual manipulation show how the model expresses short-horizon motion in the context of the observed scene and task instruction.

\begingroup
\setlength{\intextsep}{6pt}
\captionsetup{skip=5pt}
\begin{figure}[H]
    \centering
    \includegraphics[width=0.98\linewidth]{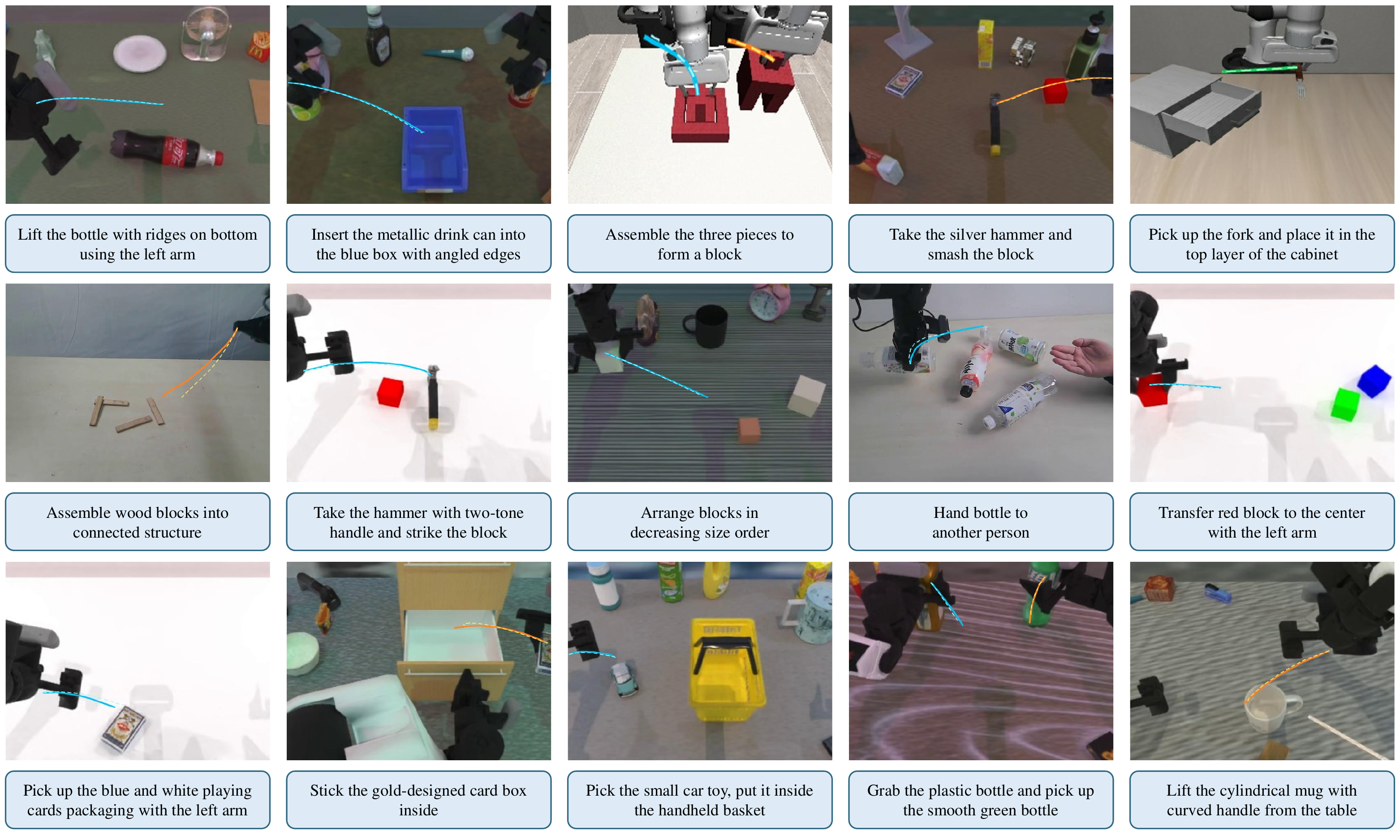}
    \caption{\textbf{Image-plane visualization of action trajectory predictions on held-out episodes.}
    Predictions are generated by the post-trained model.
    Given a task instruction, the current image, and a preceding action
    chunk, PhysBrain~1.5 predicts the next 16 action steps for the requested
    wrist or wrists. Predicted and ground-truth end-effector positions
    are projected into the image plane and overlaid on the current
    observation: solid lines denote ground-truth trajectories, and dashed
    lines denote predictions. The corresponding task instruction is shown
    below each example. All examples are drawn from test episodes unseen
    during training. These visualizations compare predicted trajectories
    with ground truth rather than report executed robot rollouts.}
    \label{fig:action_prediction_2d}
\end{figure}

\begin{figure}[H]
    \centering
    \includegraphics[width=0.98\linewidth]{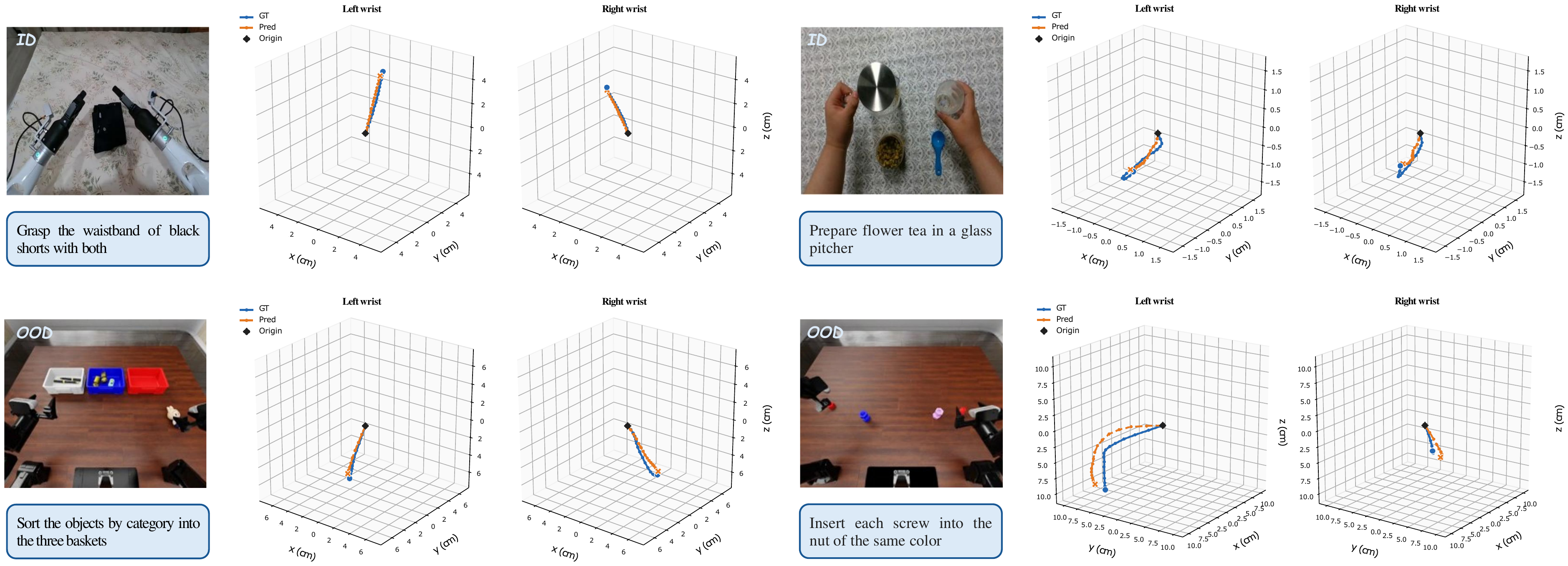}
    \caption{\textbf{Three-dimensional visualization of action trajectory predictions in ID and OOD test scenes.}
    Predictions are generated by the post-trained model.
    Each example shows the current observation and task instruction,
    alongside ground-truth (blue) and predicted (orange) trajectories
    for the left and right wrists. Black diamonds mark the trajectory
    origins, and axes are measured in centimeters. The $x$, $y$, and $z$
    axes follow the coordinate conventions of the original robot action
    data for each example; coordinate systems are not aligned across
    sources. The top row presents in-distribution (ID) examples, while
    the bottom row presents out-of-distribution (OOD) examples from
    RoboDojo, which is excluded from the model's training corpus.
    These examples provide a qualitative comparison of predicted and
    ground-truth motion within and beyond the training distribution.}
    \label{fig:action_prediction_3d}
\end{figure}
\endgroup

Figure~\ref{fig:action_prediction_3d} visualizes the predicted and reference wrist trajectories in three-dimensional coordinates for ID and OOD test scenes, using the same history-conditioned prediction protocol. The ID examples capture both near-linear motion and curved paths, while the RoboDojo examples extend the comparison to a data source excluded from training. In the OOD sorting and screw-insertion examples, the predictions reproduce the broad motion trends of the reference trajectories, although positional deviations remain visible. Each comparison retains the coordinate conventions of its source action data, without cross-source coordinate alignment. Together, these examples illustrate how PhysBrain~1.5 extends beyond scene understanding to generating structured motion continuations through its shared autoregressive interface. The evidence concerns offline trajectory prediction conditioned on observed action history, rather than closed-loop execution or full-task success.

\FloatBarrier
\subsubsection{Future Visual Prediction}

\begingroup
\setlength{\intextsep}{5pt}
\captionsetup{skip=4pt}
\begin{figure}[!t]
    \centering
    \includegraphics[width=0.98\linewidth]{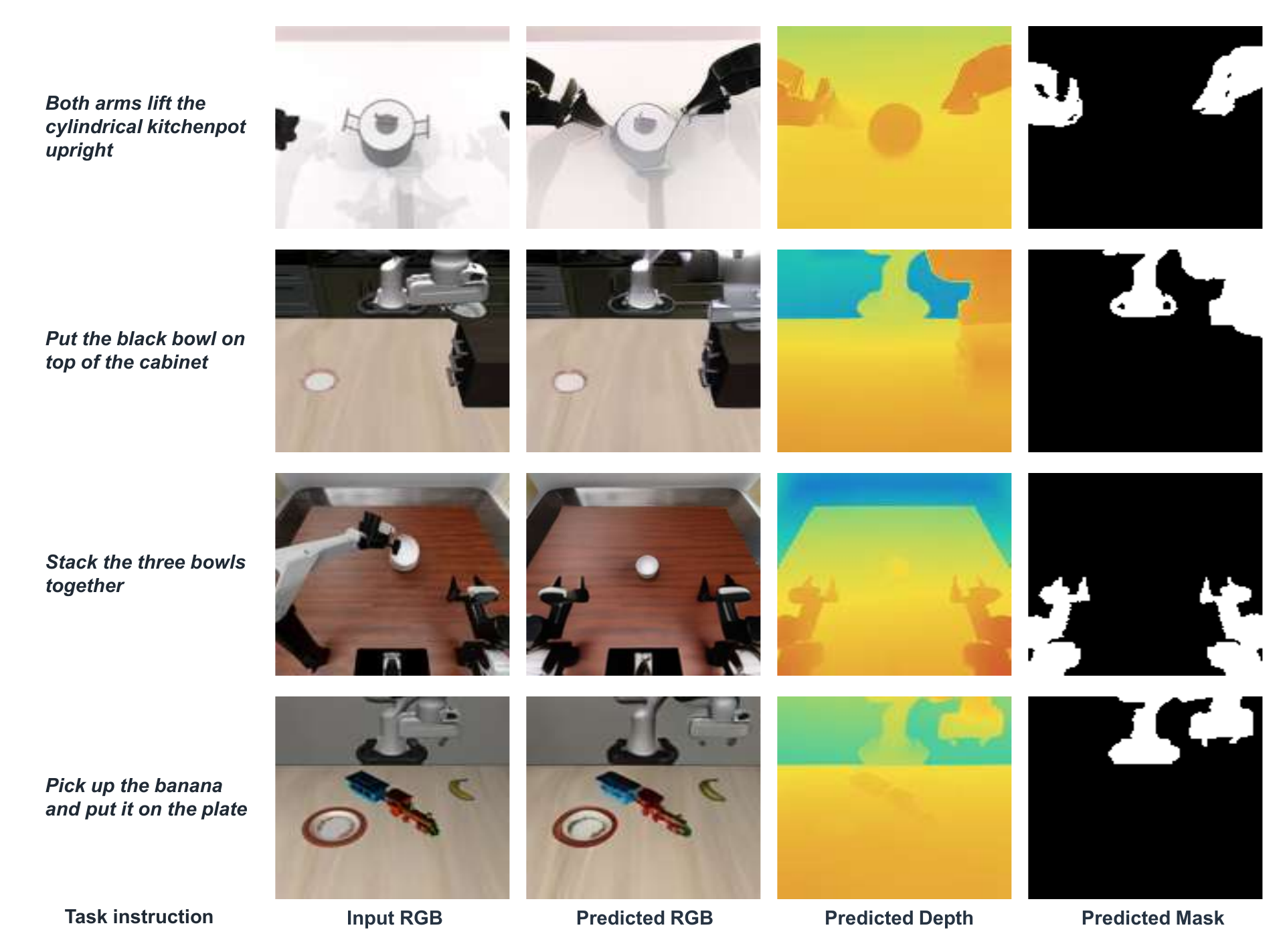}
    \caption{\textbf{Qualitative results of future visual prediction across diverse robot embodiments.} Each row presents the current RGB observation and the predicted future RGB image, depth map, and robot mask. The examples shown use a one-second prediction horizon. The model preserves the dominant scene layout and object identity while anticipating task-consistent changes. Predictions across the three modalities are spatially coherent, jointly capturing future appearance, workspace geometry, and robot occupancy.}
    \label{fig:visual_prediction}
\end{figure}
\endgroup

Given the current RGB observation and a task instruction, PhysBrain 1.5 predicts future visual states through RGB images, depth maps, and robot masks. Figure~\ref{fig:visual_prediction} presents representative examples on a one-second prediction horizon. These predictions preserve the overall scene layout while depicting plausible changes consistent with task progress across different robot embodiments and viewpoints. In the pot-lifting example, both arms move toward the pot, with their predicted positions reflected in the corresponding depth and mask outputs. More broadly, the three modalities provide spatially coherent descriptions of the future appearance, workspace geometry, and robot configuration.

These examples demonstrate how PhysBrain 1.5 extends beyond understanding the current scene to predicting task-relevant future states. Through a shared vocabulary and decoder, the model expresses these predictions in complementary visual modalities, illustrating its capacity to jointly model how the environment and robot may evolve during physical interaction.

\FloatBarrier

\section{Conclusion}
\label{sec:conclusion}

In this report, we introduced PhysBrain~1.5, guided by the physical loop in
which an agent observes its environment, acts upon it, and uses the resulting
changes to inform subsequent behavior. The model supports embodied
understanding, action generation, and future-state prediction through a shared
autoregressive backbone, representing language responses, end-effector
trajectories, and multimodal visual states as discrete tokens under a common
next-token prediction objective. Physical-aware pre-training derives its
embodied supervision entirely from human interaction videos, alongside general
language and vision--language data. Supervised fine-tuning then incorporates
human, real-robot, and simulated interaction data.

On 28 embodied understanding benchmarks, our 8B model achieves an average
score of 72.5, setting a new open-source state of the art and
approaching leading proprietary systems, while
retaining broad multimodal understanding capabilities. Qualitative results
further demonstrate end-effector trajectory prediction and spatially coherent
predictions of future RGB images, depth maps, and robot masks. These results
support learning from interaction experience as an effective route toward
developing the capabilities needed for the physical loop. We will continue scaling the volume of training data, expanding modality coverage, and incorporating richer interaction experience.

\clearpage

{
	\bibliographystyle{unsrtnat}
	\bibliography{ref}
}

\clearpage

\setcounter{section}{0}
\renewcommand{\thesection}{\Alph{section}}
\renewcommand{\thesubsection}{\thesection.\arabic{subsection}}
\setcounter{subsection}{0}
\renewcommand{\thesubsubsection}{\thesubsection.\arabic{subsubsection}}
\renewcommand{\theHsection}{app.\Alph{section}}
\renewcommand{\theHsubsection}{app.\Alph{section}.\arabic{subsection}}
\renewcommand{\theHsubsubsection}{app.\Alph{section}.\arabic{subsection}.\arabic{subsubsection}}
\setcounter{tocdepth}{1}

\section{Contributor}
\label{app:contributors}
The contributors are listed alphabetically by family name:

Yu Bin, Haipeng Cao, Zheng Chang, Kai Chen, Youning Chen, Kailin Deng, Yichao Du, Xiaotong Fu, Haoyang Ge, Yunlong Guo, Chenliu Hao, Jiyan He, Xuguo He, Yakun Hou, Kai Hu, Cong Huang, Tuopusen Huang, Yu Huang, Hong Li, Peize Li, Shijie Lian, Xiaopeng Lin, Yun Lin, Haibao Liu, Haochen Liu, Qiuzhi Liu, Shengcai Liu, Zhiqiang Liu, Tao Luo, Peng Ren, Shuo Ren, Chaoyi Ruan, Zhaolong Shen, Yukun Shi, Qiyuan Su, Yuxuan Tian, Yining Wang, Changti Wu, Hao Wu, Xueyin Xu, Ruoqi Yang, Zhaoyang Yang, Hang Yuan, Zhaoyang Zeng, Hanwen Zhang, Ruimeng Zhang, Yao Zhang, Yibo Zhang, Yuxiang Zhang, Zhirui Zhang, Ziyi Zhang, Zubin Zheng, Zishen Zhuang

\section{Evaluation Protocols and Visualizations}
\label{app:eval-details}

\subsection{Embodied Evaluation Protocols and Metric Standardization}
\label{app:embodied-eval-protocols}

\subsubsection{Framework and Protocol Alignment}

EmbodiedEvalKit\footnote{\url{https://github.com/pickxiguapi/EmbodiedEvalKit}} is a unified evaluation framework that integrates a broad collection of benchmarks for embodied perception, spatial understanding, reasoning, and planning under consistent model interfaces and evaluation workflows. We directly reuse its benchmark-specific input protocols, prompting templates, and evaluation metrics for ERQA, EgoPlan-Bench2, SAT, ShareRobot-Trajectory, VABench-Visual-Trace, BLINK, CV-Bench, VSI-Bench, EmbSpatial, COSMOS, RoboVQA, and VLABench. For RoboVQA, we retain the original question formulation and evaluation metric while extending the visual input from the eight-frame setting used in EmbodiedEvalKit to the full set of 16 frames provided by the official benchmark, thereby preserving more of the available temporal context. For ShareRobot-Trajectory and VABench-Visual-Trace, we follow EmbodiedEvalKit to compute the normalized RMSE (NRMSE) and report $100-\mathrm{NRMSE}$ as the final score, ensuring that higher values consistently indicate better performance across all benchmarks. Except for the RoboVQA extension described above, we retain the default visual-input configurations of EmbodiedEvalKit for all benchmarks inherited from the framework, including the source-image resolution and video-frame sampling protocol. 

\subsubsection{Unified Point-Localization Metric}

For the ten point-localization and grounding benchmarks---Part-Affordance-2K, PIOBench S1 \& S2, PixMo-Points, PointBench, RefSpatial-Bench, RoboAfford, RoboRefit, RoboSpatial-Home, VABench-Point, and Where2Place---we follow the input protocols and prompting templates of EmbodiedEvalKit but adopt a unified evaluation metric. We observe that the original implementations employ heterogeneous criteria: most measure only the fraction of predicted points that fall inside valid target regions, which is equivalent to precision, whereas others measure only the coverage of ground-truth regions, which is equivalent to recall. Either criterion alone provides an incomplete assessment, as it does not simultaneously penalize invalid predictions and missed targets. We therefore evaluate all point-localization tasks using a micro-averaged F1 score that jointly measures localization correctness and target coverage. Whenever instance-level annotations are available, we determine point validity using the corresponding target masks. In particular, the original EmbodiedEvalKit implementation of RoboRefit evaluates predictions against bounding-box annotations, whereas we use corrected instance masks to avoid accepting points that lie inside a coarse bounding box but outside the actual target object. Specifically, for sample $i$, let $\widehat{\mathcal{P}}_i$ be the collection of predicted points and $\mathcal{G}_i$ be the set of ground-truth target regions. We define the number of valid predicted points as
\begin{equation}
H_i =
\sum_{p \in \widehat{\mathcal{P}}_i}
\mathbf{1}\!\left[
\exists\, g \in \mathcal{G}_i
\ \text{s.t.}\ 
p \in g
\right],
\end{equation}
where a prediction is considered valid whenever it falls inside at least one ground-truth target region. To measure target coverage, we construct a set of valid one-to-one matches $\mathcal{M}_i \subseteq \widehat{\mathcal{P}}_i \times \mathcal{G}_i$, where $(p,g)\in\mathcal{M}_i$ only if $p\in g$, and each predicted point and target region can participate in at most one match. The number of covered targets is therefore
\begin{equation}
M_i = \left|\mathcal{M}_i\right|.
\end{equation}
For multi-instance samples in PixMo-Points, we first obtain a one-to-one assignment between predicted points and annotated target points using distance-based Hungarian matching, and then validate each assigned prediction against the corresponding instance mask. We aggregate the corresponding counts over the complete benchmark and compute micro-averaged precision and recall as
\begin{equation}
P =
\frac{\sum_i H_i}
     {\sum_i \left|\widehat{\mathcal{P}}_i\right|},
\qquad
R =
\frac{\sum_i M_i}
     {\sum_i \left|\mathcal{G}_i\right|}.
\end{equation}
The final point-localization score is their harmonic mean:
\begin{equation}
F_1 =
\frac{2PR}{P+R}.
\end{equation}
This formulation accepts any point located within a valid target region without requiring it to reproduce an arbitrary annotated coordinate, while penalizing both predictions outside valid regions and ground-truth targets that are not covered. For samples containing no valid target, an explicit no-object prediction is treated as correct, whereas any predicted coordinate is treated as invalid. For PointBench samples that explicitly require a particular number of points, we additionally retain the benchmark's exact-count constraint: a prediction with an incorrect number of points contributes no valid matches. For RoboSpatial-Home, which contains both point-localization and binary question types, we apply the unified F1 metric to the point-localization subset and retain accuracy for the binary subset. Their combined score is computed by sample-count-weighted averaging:
\begin{equation}
S_{\mathrm{RoboSpatial}}
=
\frac{
N_{\mathrm{point}}F_{1,\mathrm{point}}
+
N_{\mathrm{binary}}A_{\mathrm{binary}}
}{
N_{\mathrm{point}}+N_{\mathrm{binary}}
},
\end{equation}
where $N_{\mathrm{point}}$ and $N_{\mathrm{binary}}$ denote the numbers of point-localization and binary samples, respectively, and $A_{\mathrm{binary}}$ denotes the accuracy of the binary subset.

\subsubsection{Additional Benchmark Integrations}
The remaining six benchmarks are not included in the original EmbodiedEvalKit and are integrated from their official releases. For ERQA-PLUS, ViewSpatial-Bench, MindCube, and MMSI-Bench, we preserve the released questions, answer choices, visual inputs, and their original ordering, and report top-1 accuracy. Specifically, we evaluate the complete 1,766-question release of ERQA-PLUS, all 5,712 questions in ViewSpatial-Bench, the 1,050-question MindCube TinyBench split under its raw-QA setting without additional cognitive-map or reasoning scaffolds, and the standard 1,000-question MMSI-Bench protocol without circular answer-choice augmentation. For these multiple-choice benchmarks, we retain the official task formulations while expressing the final answer instruction in a shared answer-only format. For 3DSRBench, we use the officially released circular TSV containing both FlipEval-augmented questions and circularly permuted answer choices. Predictions are grouped by their base question, and a group is considered correct only when all of its circular variants are answered correctly, following the official aggregation procedure. For Q-Spatial-Bench, we evaluate the 101 samples in the QSpatial-plus split and follow its official success-rate metric, which normalizes the predicted and ground-truth distances to a common unit and considers a prediction successful when it falls within a factor of two of the ground-truth distance.

\subsection{General Multimodal Evaluation Protocol}
\label{app:general-mm-protocol}

The general multimodal suite covers complementary task types. MME measures
visual perception and cognition, MMStar evaluates multimodal reasoning, and
RealWorldQA tests visual understanding of real-world scenes. VideoMME and
MVBench evaluate temporal and spatiotemporal reasoning in videos. POPE probes
object hallucination, AI2D and ChartQA assess reasoning over diagrams and
charts, and DocVQA and TextVQA test understanding of documents and text in
natural images. V* emphasizes fine-grained recognition in high-resolution
images, while ScreenSpot evaluates visual GUI grounding.

\subsubsection{Image Evaluation}

The Qwen-aligned image profile uses non-thinking inference, a maximum of
32,768 generated tokens, temperature $0.7$, top-$p$ $0.8$, top-$k=20$,
repetition penalty $1.0$, presence penalty $1.5$, and seed 3407. The maximum
image area is 4,014,080 pixels. The minimum image area is 602,112 pixels
for RealWorldQA and 200,704 pixels for the remaining image tasks.

ScreenSpot uses Qwen's computer-use tool-call format with coordinates on
a fixed 0--1000 grid. RealWorldQA uses rule-based extraction with a fixed
local Qwen3 8B model fallback shared by both models; this differs
from Qwen's remote fallback judge. DocVQA and TextVQA use their validation
splits.

\subsubsection{Video Evaluation}

\paragraph{Video inputs and temporal sampling}
We use the Qwen3-VL video preprocessing pipeline implemented in
\texttt{qwen-vl-utils}, with a target sampling rate of 2 FPS and a maximum
of 128 frames per clip for VideoMME and 32 for MVBench, using the same
benchmark-specific limits for both models. The target frame count is determined from the
clip duration, with a default minimum of four frames subject to the
available source frames, and is rounded down to a multiple of two to
match the model's temporal factor. Frames are then sampled uniformly
across the full input clip using evenly spaced frame indices rounded
to the nearest integer. Thus, short clips may contain fewer sampled
frames than the corresponding benchmark limit, while the effective sampling rate decreases for
long clips once the frame limit is reached. Frame indices and source
frame-rate metadata are passed to the model's processor to preserve
temporal information.

\paragraph{Spatial preprocessing}
Sampled frames are resized with bicubic interpolation while preserving
their aspect ratio, with height and width aligned to multiples of 32
for the 16-pixel patches and spatial merging factor of two. The
configured minimum frame area is 200,704 pixels. Video preprocessing
applies its own pixel budget: the default per-frame ceiling is
$768\times32^2=786{,}432$ pixels, further constrained by the total video
budget. The resulting resolution therefore depends on the source
aspect ratio and the video budget. Frames are resized once by
\texttt{qwen-vl-utils}; subsequent processor resizing is disabled.

\paragraph{Task protocols and scoring}
VideoMME is evaluated on its complete test split of 2,700 questions,
covering short, medium, and long videos, and is scored by overall
multiple-choice accuracy. We use the version without additional
subtitles or audio input. MVBench covers 20 sub-tasks and 4,000
evaluation examples, using the video clips supplied by its video
release. The reported MVBench score is the unweighted mean of the
accuracies across its 20 sub-tasks. Both benchmarks provide a question and answer options and
request an option-letter response, which is scored using the
benchmark-specific answer-extraction rules without an external LLM
judge.

\begin{figure}[!t]
    \centering
    \includegraphics[width=1.0\linewidth]{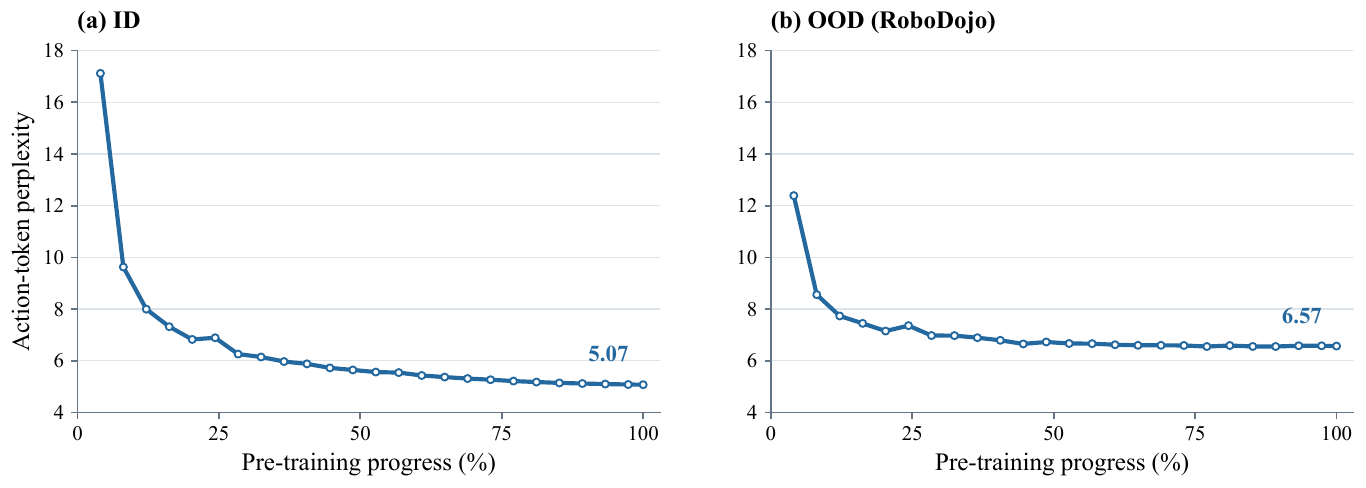}
    \caption{\textbf{Action-token perplexity during pre-training.}
    The PhysBrain 1.5 model is evaluated on ID test data and OOD data from RoboDojo, which is excluded from training. ID test episodes are disjoint from training episodes.}
    \label{fig:action_ppl}
\end{figure}

\subsection{Action Perplexity}
Figure~\ref{fig:action_ppl} tracks action-token perplexity of the 8B model during pre-training on ID test data and the held-out RoboDojo corpus. The ID test set shares data sources with training but contains disjoint episodes. All RoboDojo test examples include ground-truth action history, and RoboDojo data are excluded from training. From the first evaluated checkpoint to the last, perplexity decreases from 17.12 to 5.07 on ID data and from 12.39 to 6.57 on RoboDojo. Despite local fluctuations, RoboDojo perplexity also follows an overall downward trend as pre-training progresses, indicating improved prediction of reference action tokens on a data source excluded from training.

\subsection{Qualitative Visualizations}
\label{app:qualitative_results}
\begin{figure*}[h]
    \centering

    \includegraphics[
        width=0.32\textwidth
    ]{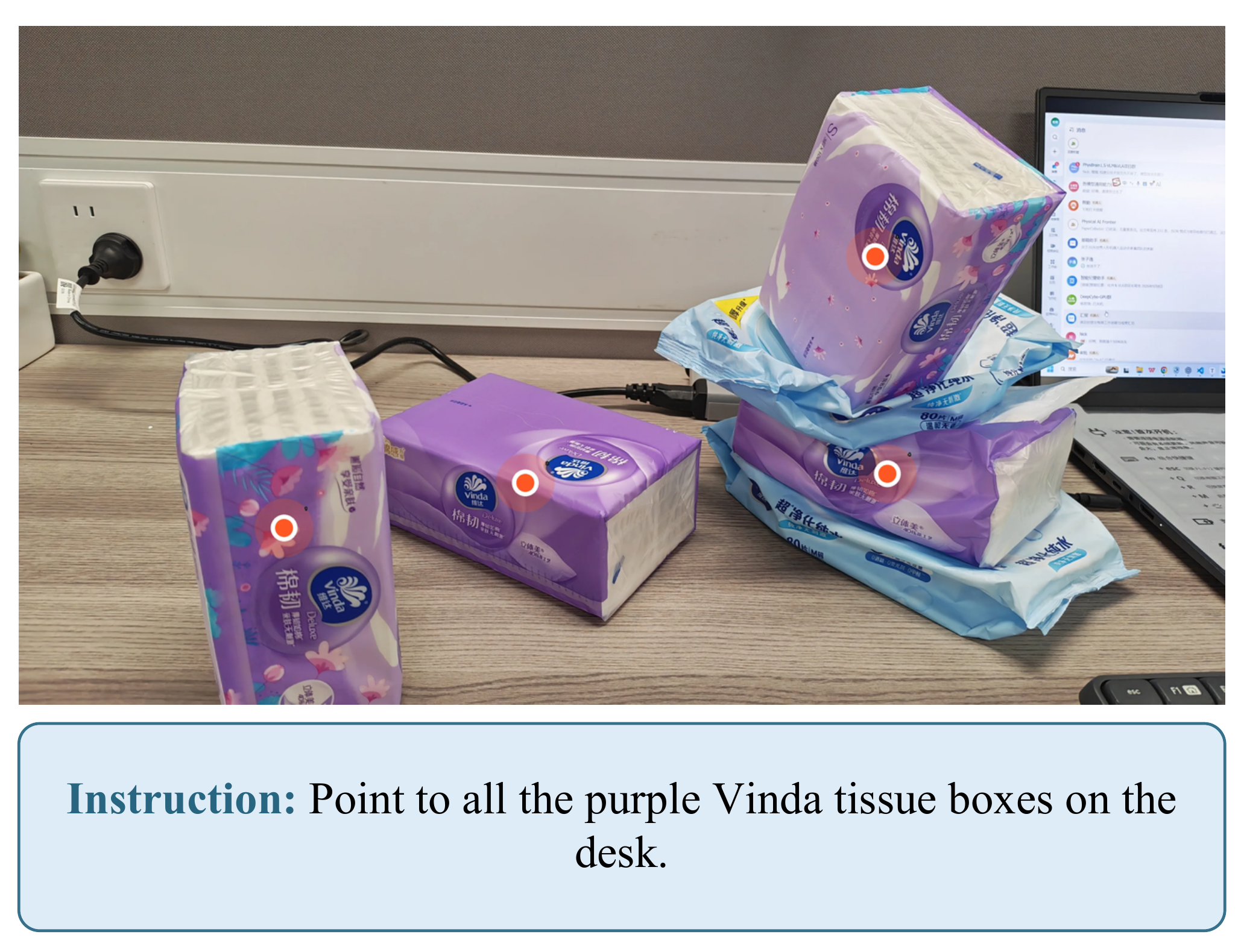}
    \hfill
    \includegraphics[
        width=0.32\textwidth
    ]{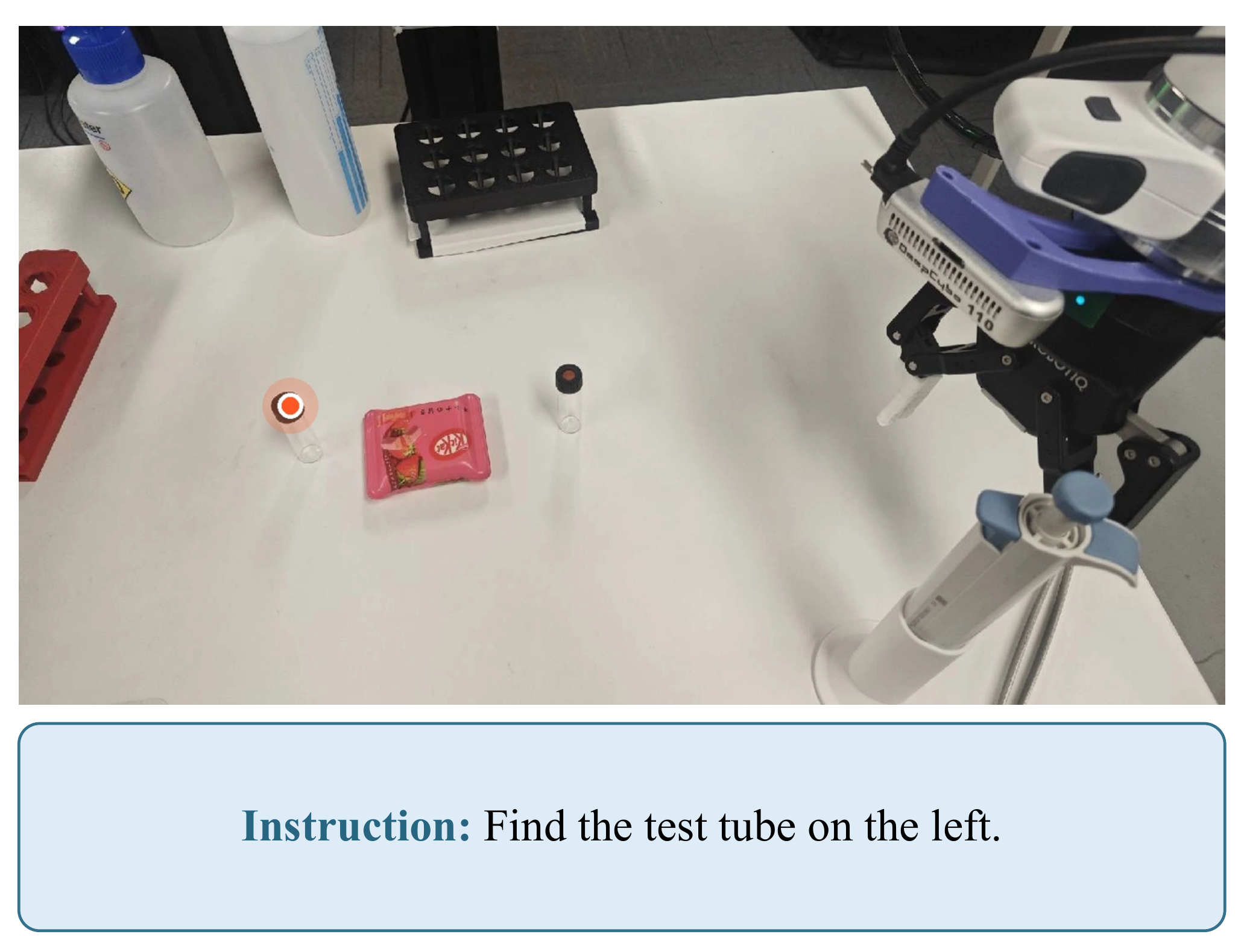}
    \hfill
    \includegraphics[
        width=0.32\textwidth
    ]{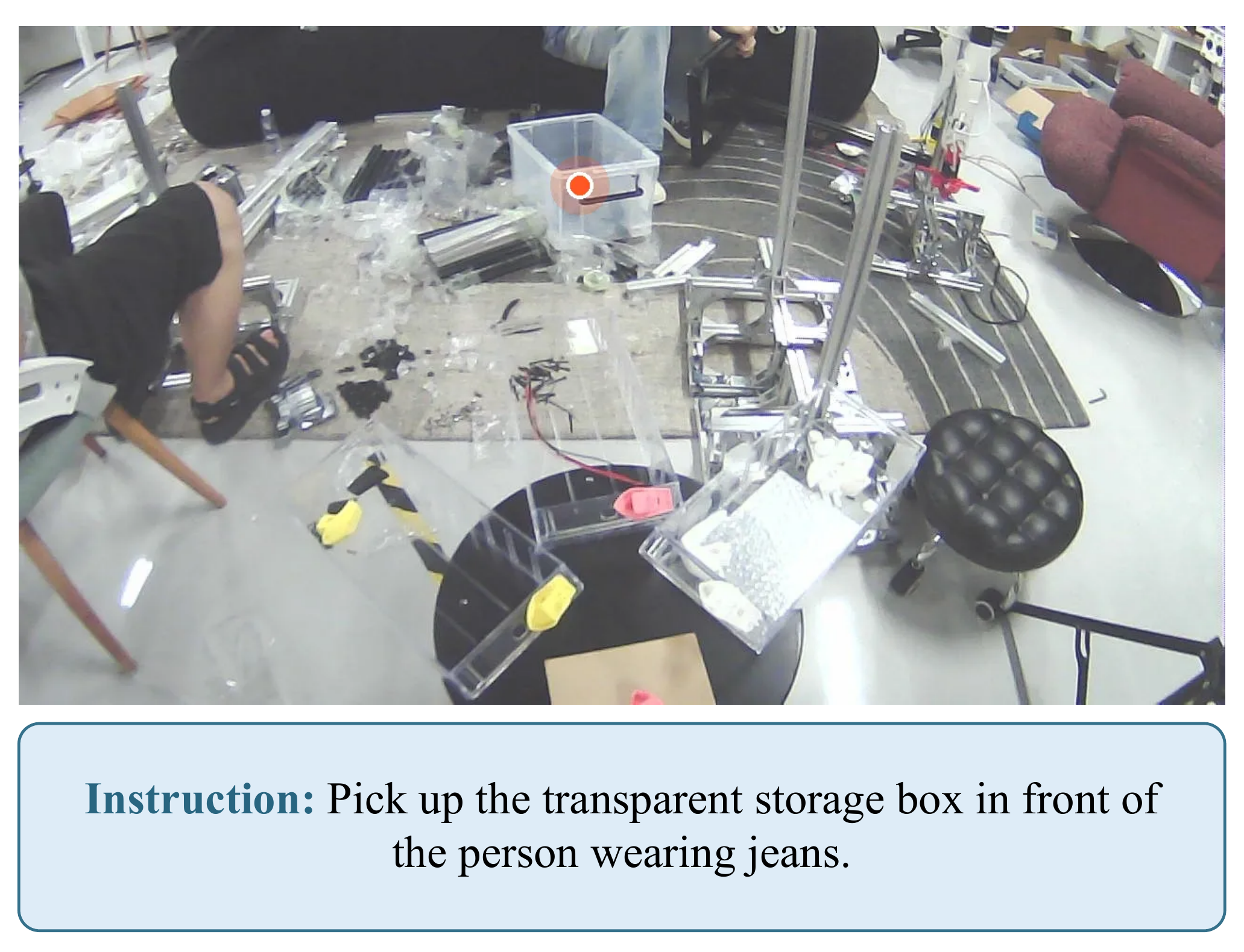}

    \par\vspace{0.8em}

    \includegraphics[
        width=0.32\textwidth
    ]{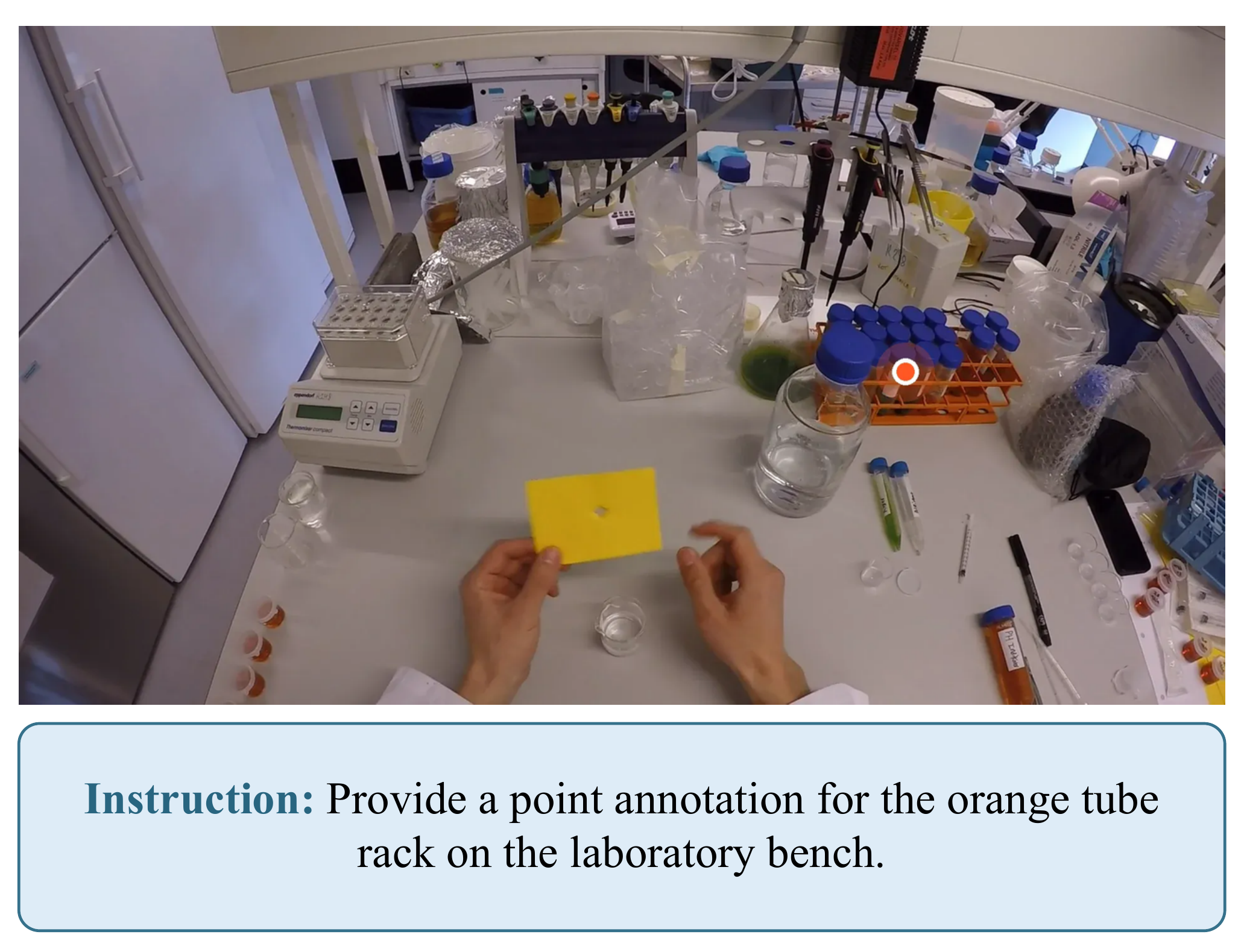}
    \hfill
    \includegraphics[
        width=0.32\textwidth
    ]{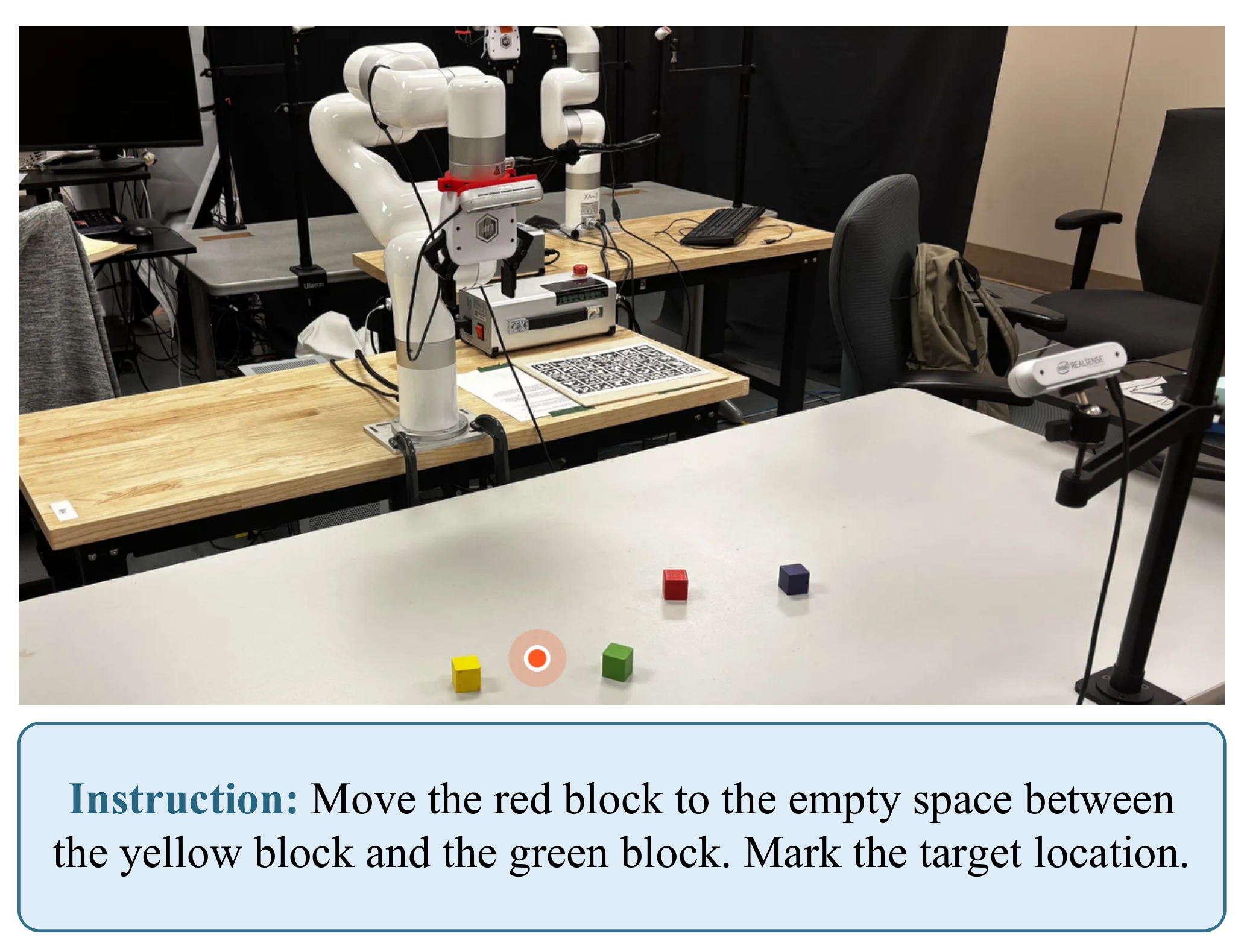}
    \hfill
    \includegraphics[
        width=0.32\textwidth
    ]{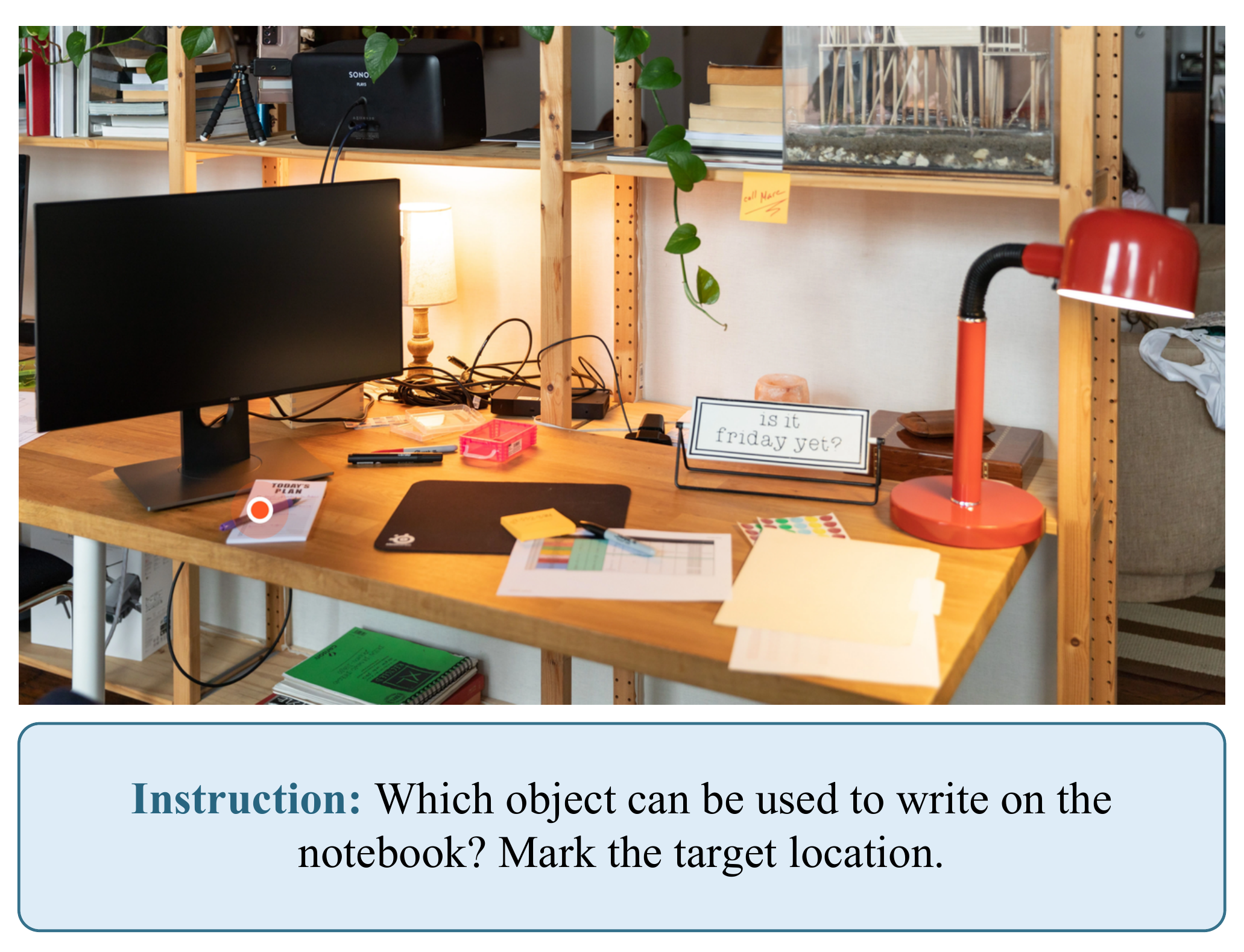}

    \caption{Examples of object pointing across diverse complex scenarios.}
    \label{fig:point-examples-vis}
\end{figure*}

\begin{figure*}[h]
    \centering

    \includegraphics[
        width=0.32\textwidth
    ]{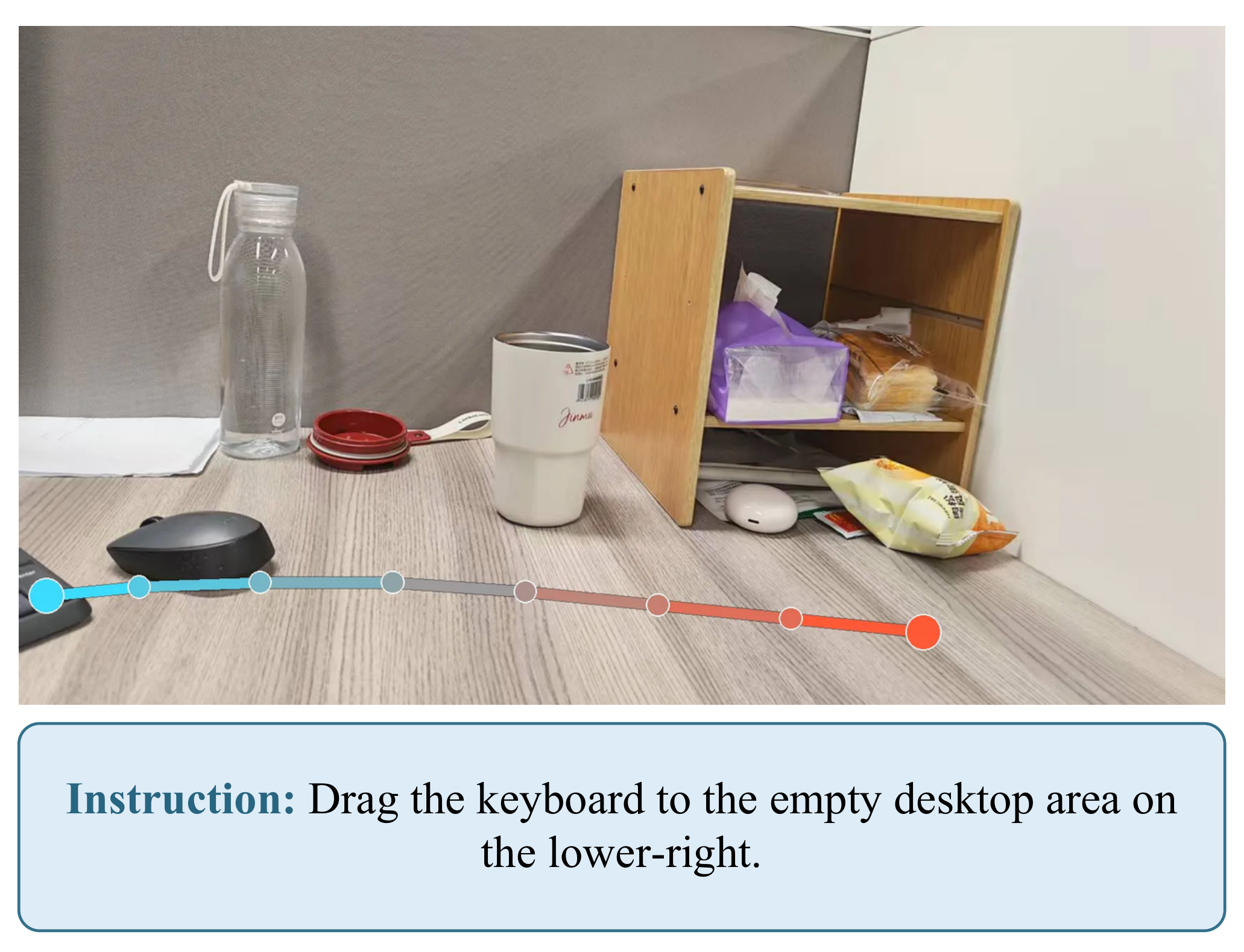}
    \hfill
    \includegraphics[
        width=0.32\textwidth
    ]{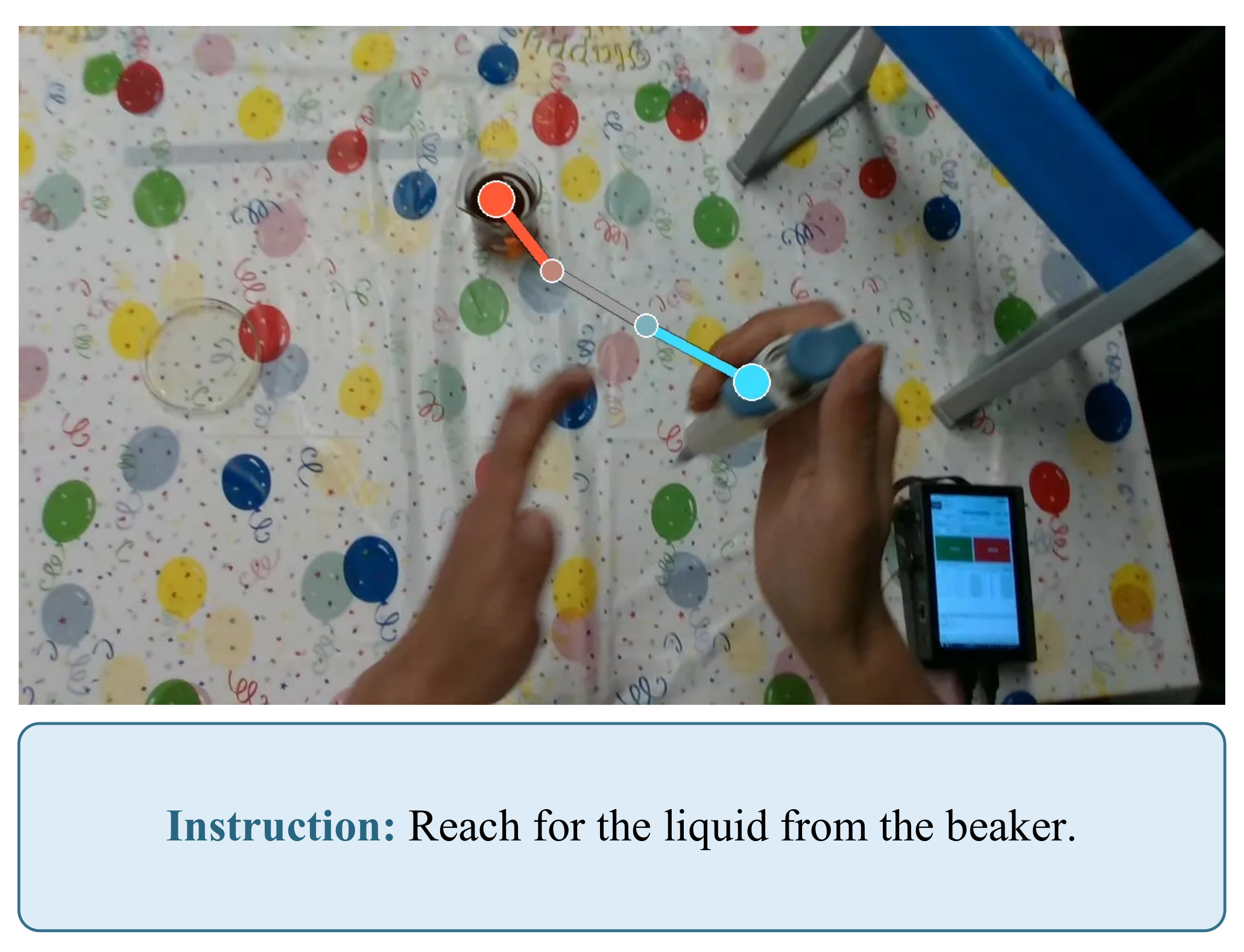}
    \hfill
    \includegraphics[
        width=0.32\textwidth
    ]{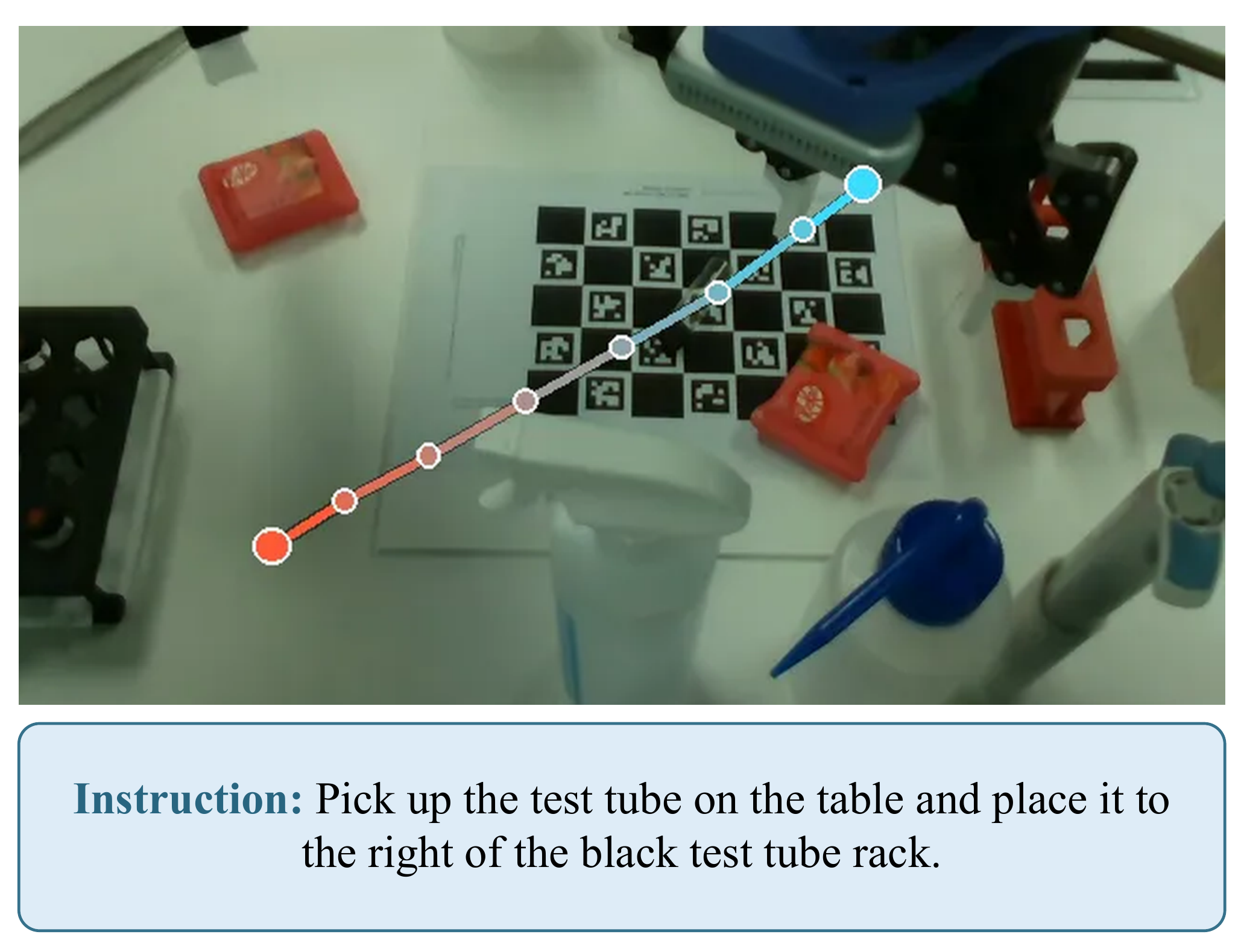}

    \par\vspace{0.8em}

    \includegraphics[
        width=0.32\textwidth
    ]{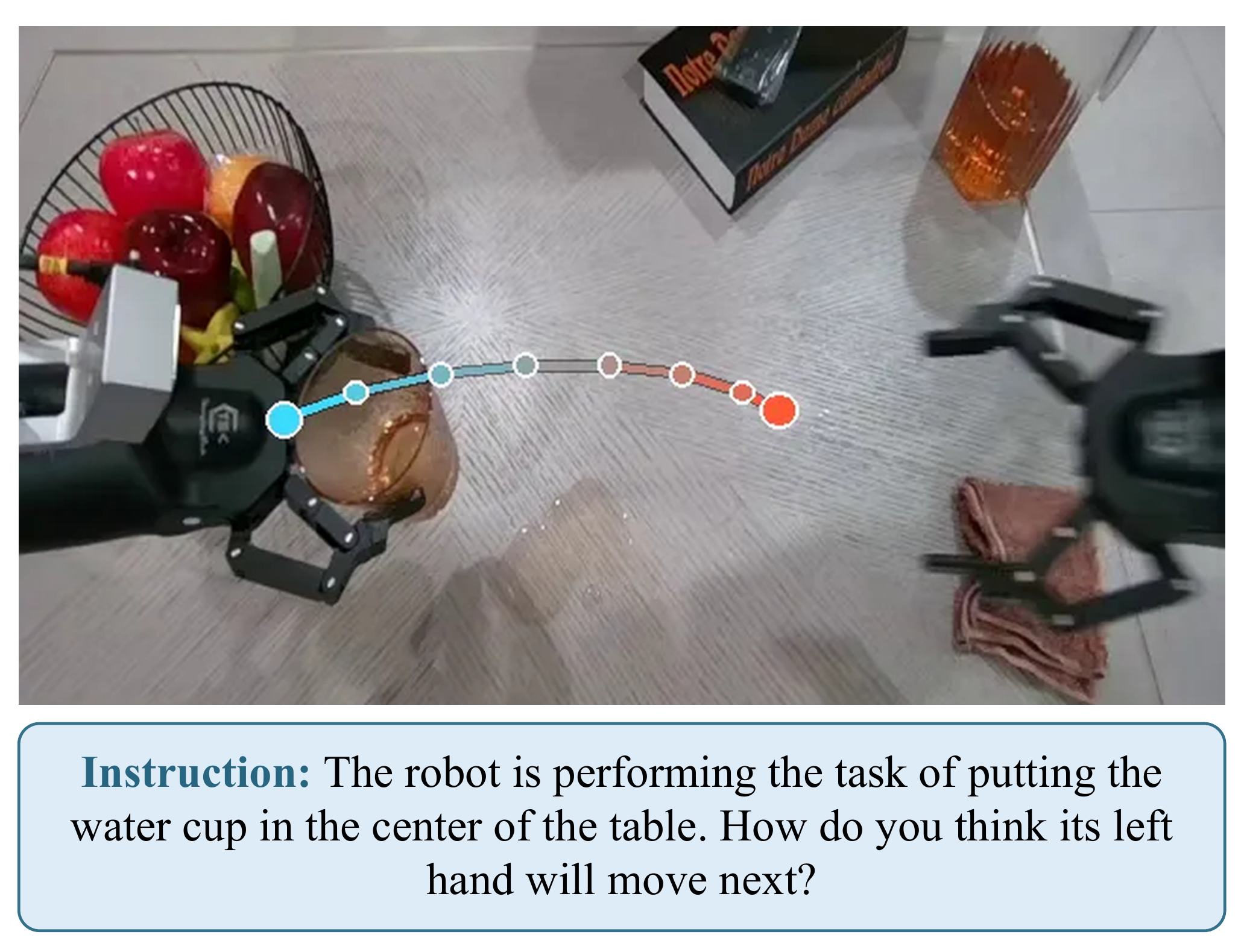}
    \hfill
    \includegraphics[
        width=0.32\textwidth
    ]{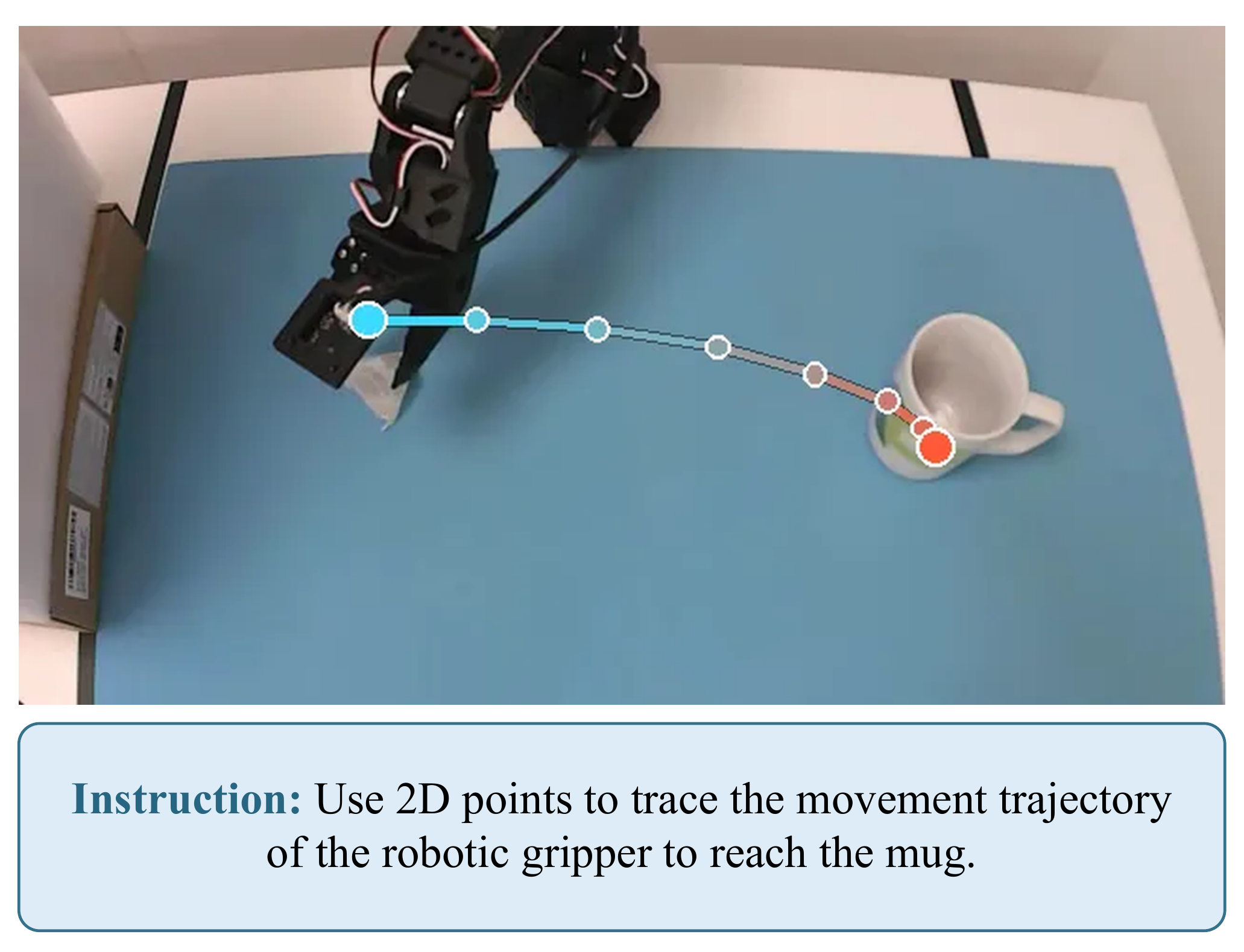}
    \hfill
    \includegraphics[
        width=0.32\textwidth
    ]{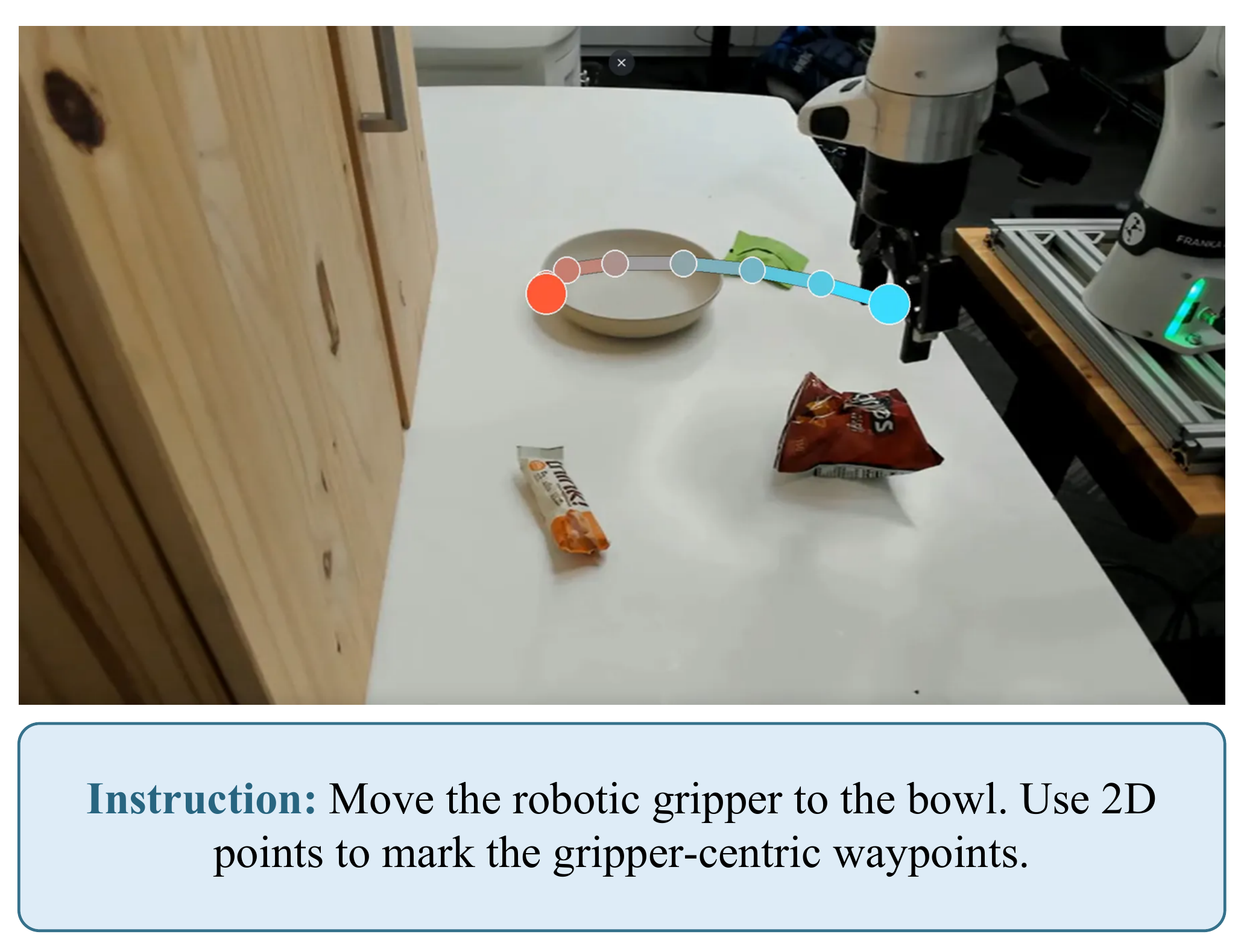}

    \caption{Examples of predicting the movement trajectories of target objects based on action instructions.}
    \label{fig:trajectory-examples-vis}
\end{figure*}

\begin{figure*}[h]
    \centering

    \includegraphics[
        width=0.32\textwidth
    ]{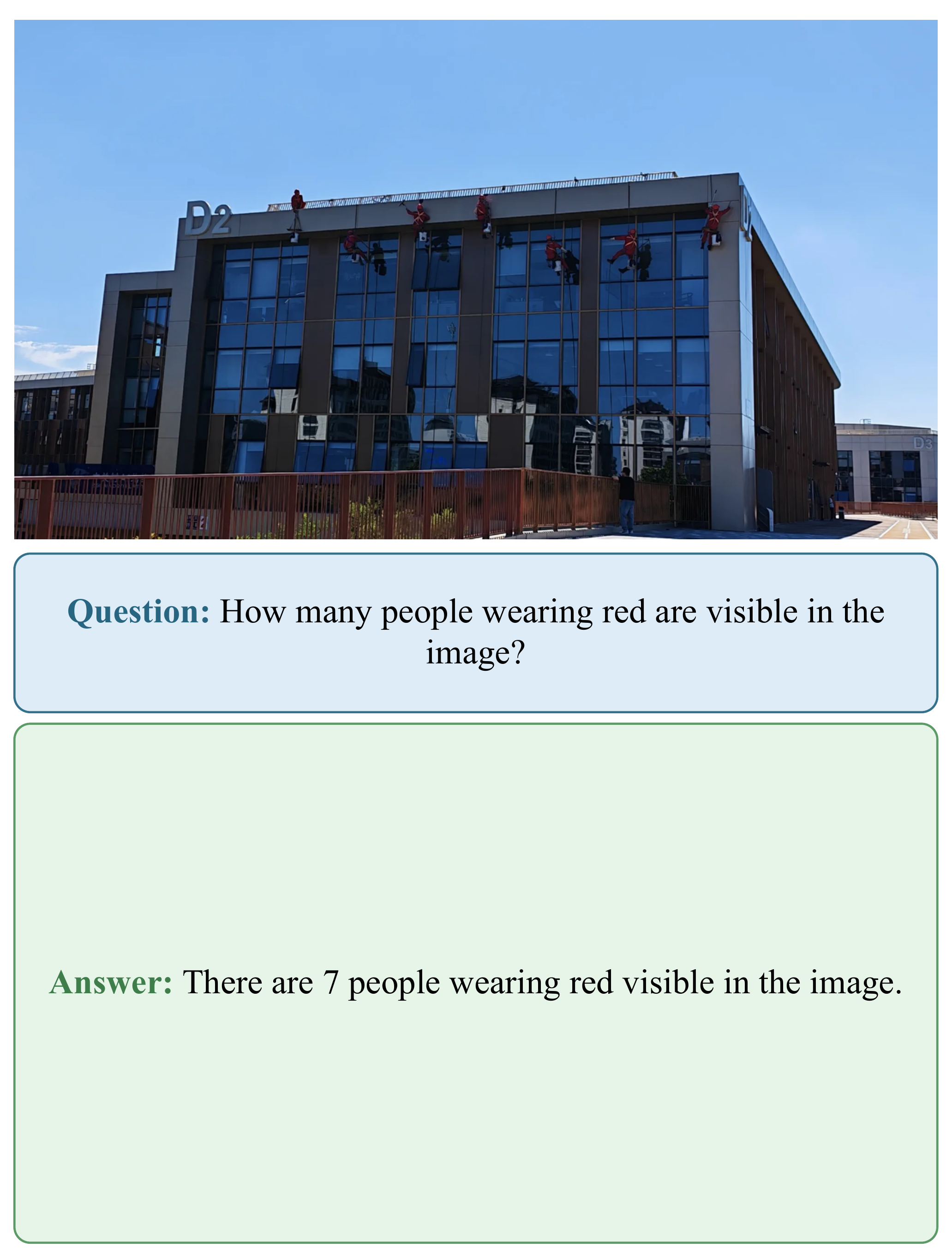}
    \hfill
    \includegraphics[
        width=0.32\textwidth
    ]{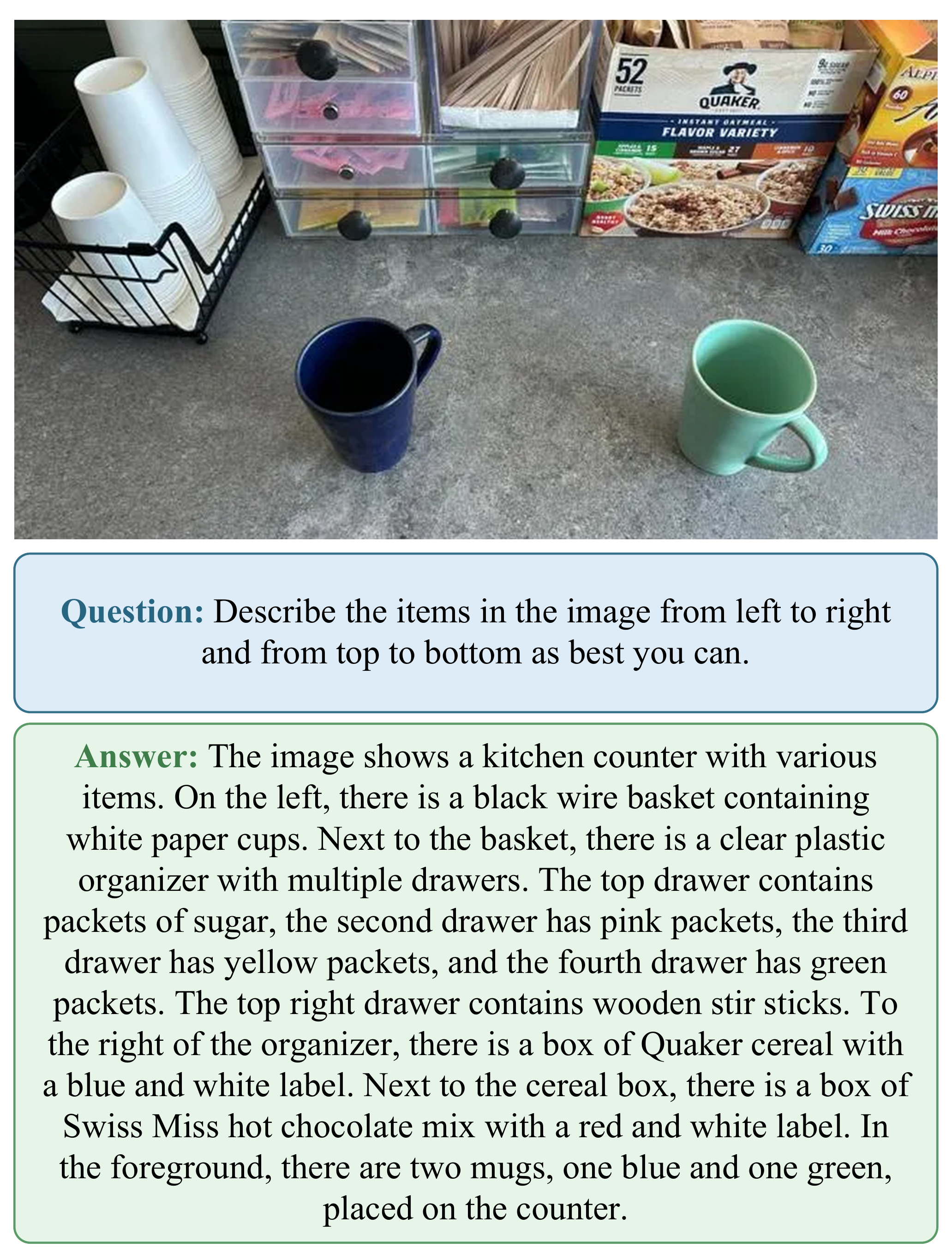}
    \hfill
    \includegraphics[
        width=0.32\textwidth
    ]{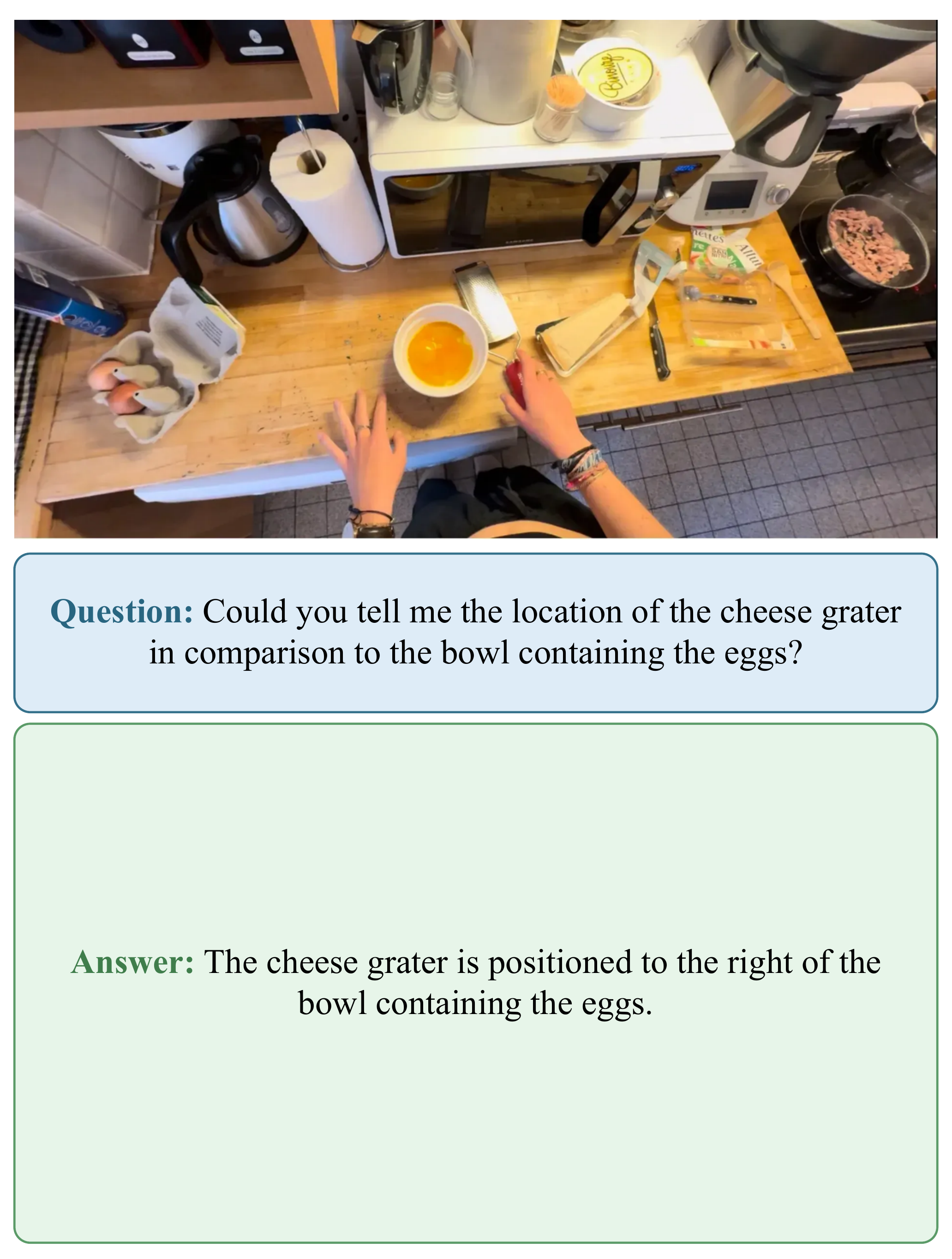}

    \par\vspace{0.8em}

    \includegraphics[
        width=0.32\textwidth
    ]{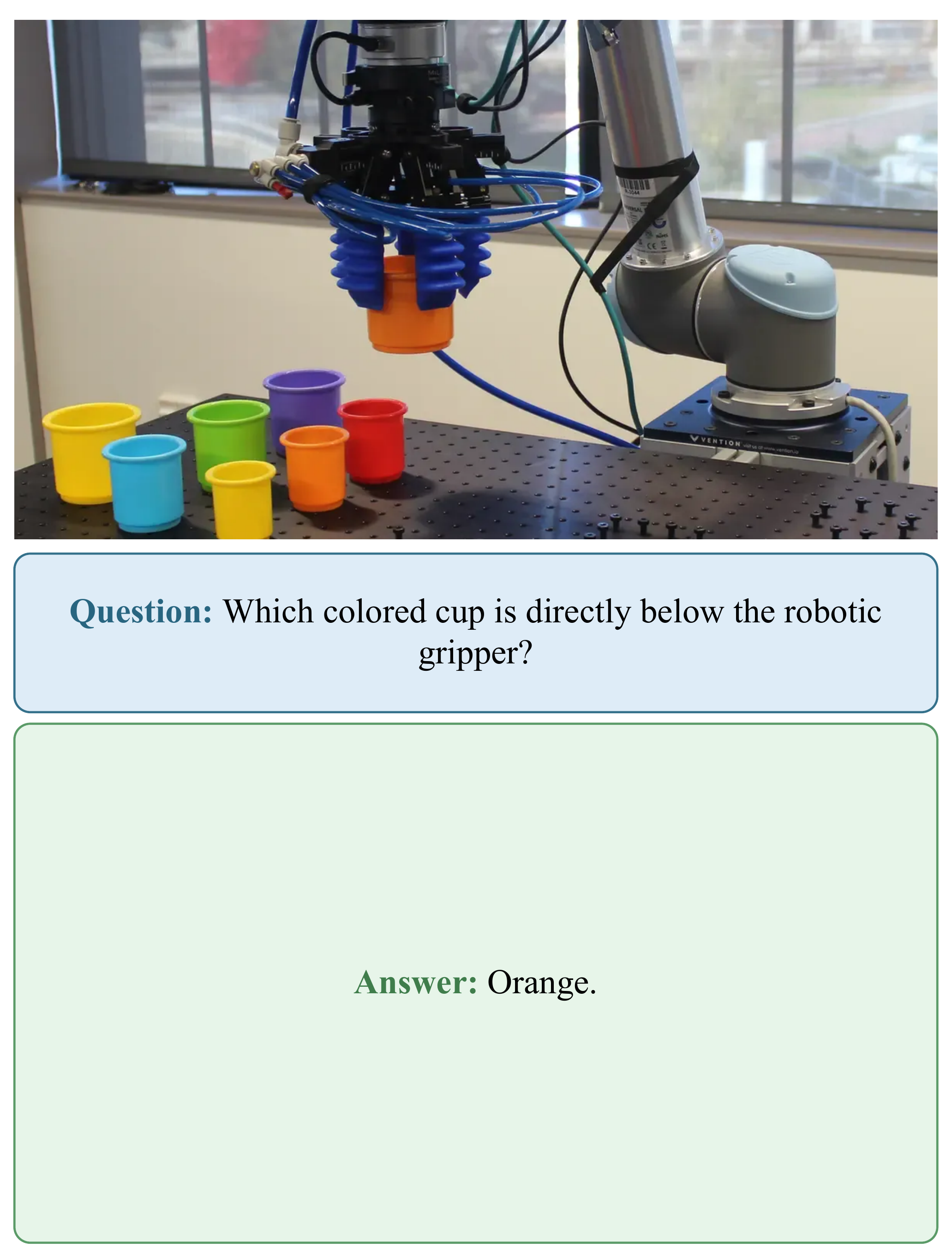}
    \hfill
    \includegraphics[
        width=0.32\textwidth
    ]{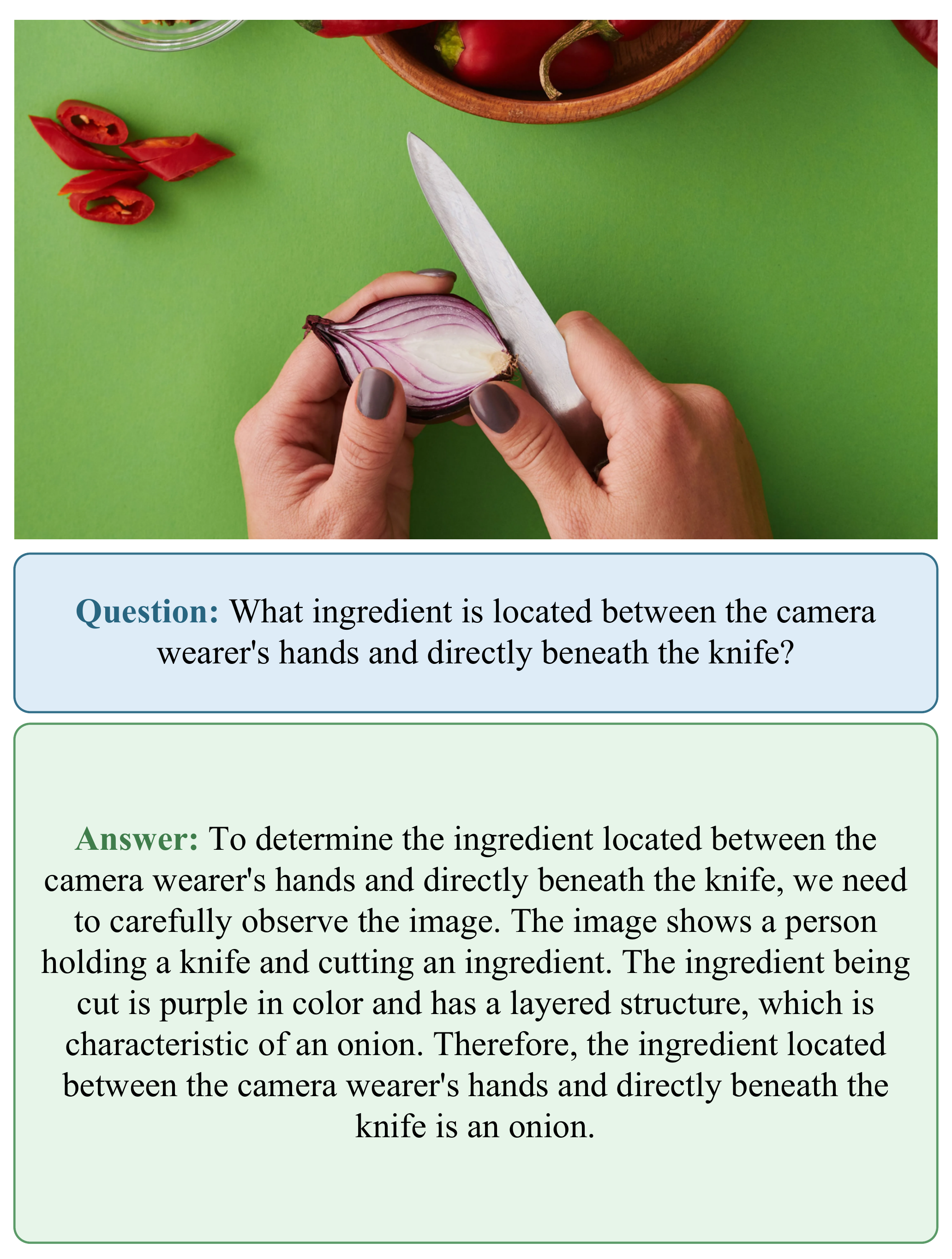}
    \hfill
    \includegraphics[
        width=0.32\textwidth
    ]{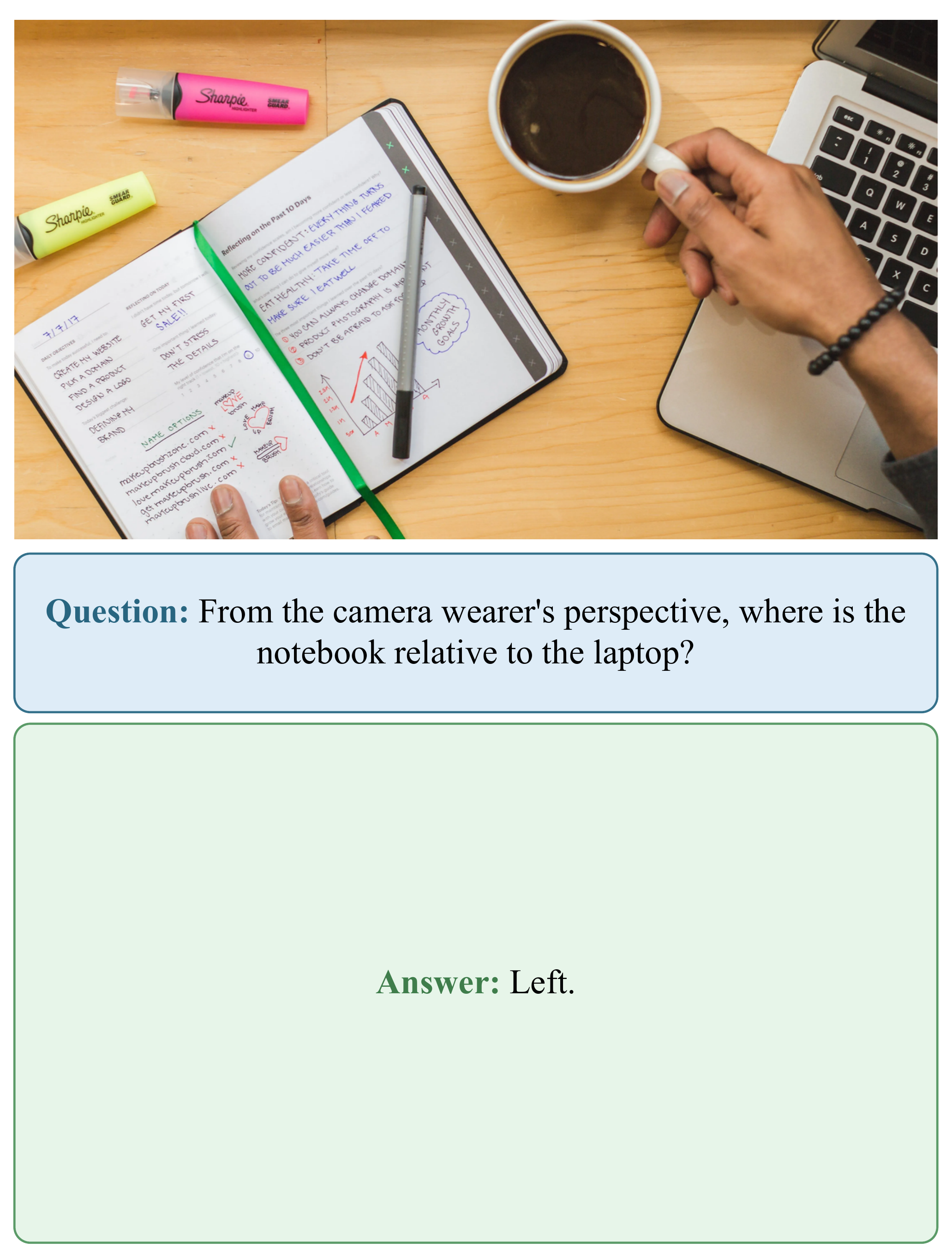}

    \caption{Examples of challenging spatial reasoning question answering.}
    \label{fig:spatial-qa-examples-vis}
\end{figure*}

\begin{figure*}[p]
    \centering
    \includegraphics[
        width=\textwidth,
        height=0.86\textheight,
        keepaspectratio
    ]{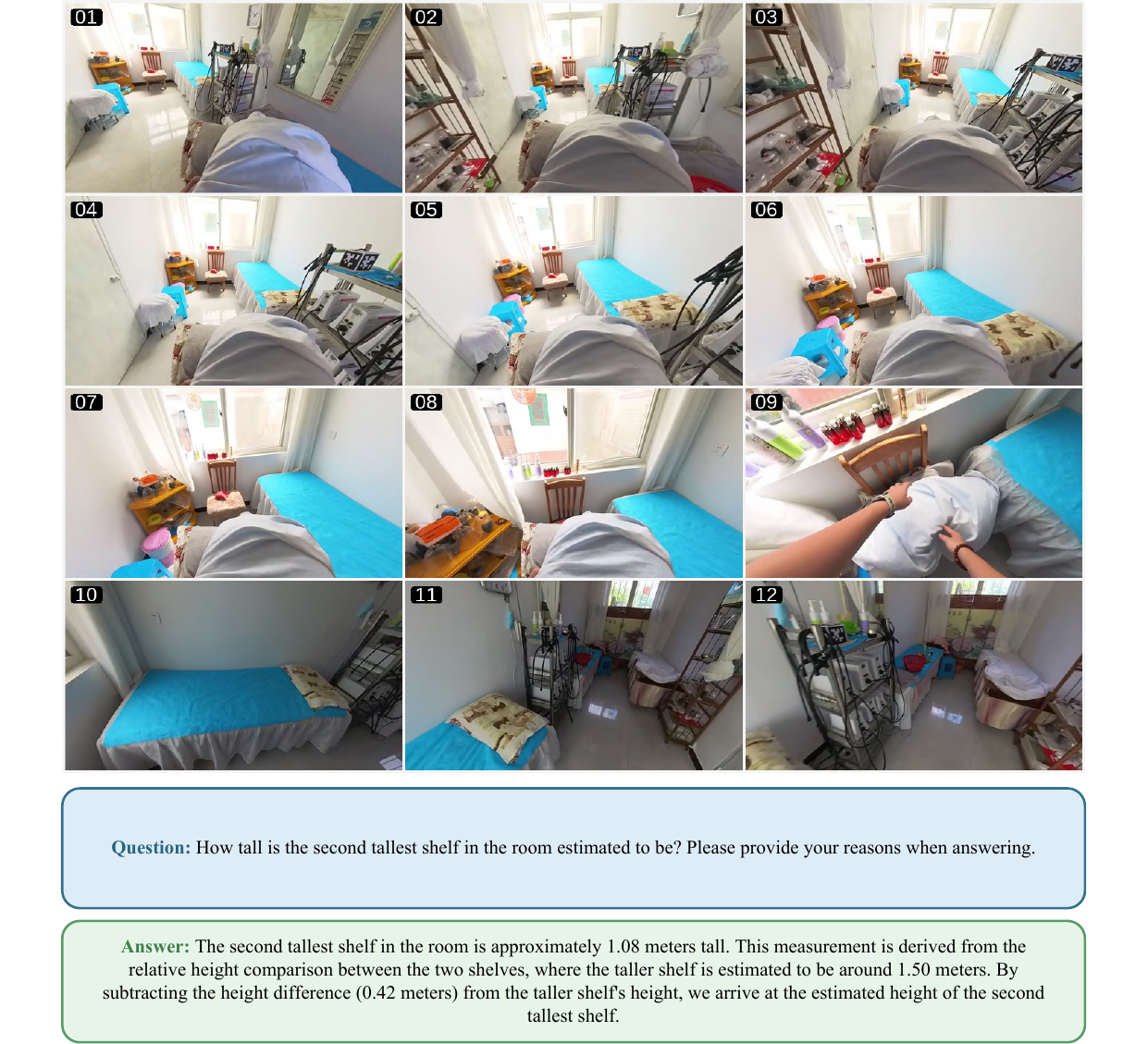}

    \caption{An example of object scale estimation in a scene with multi-view observations. }
    \label{fig:multi-image-scale-estimation}
\end{figure*}

\begin{figure*}[p]
    \centering
    \includegraphics[
        width=\textwidth,
        height=0.86\textheight,
        keepaspectratio
    ]{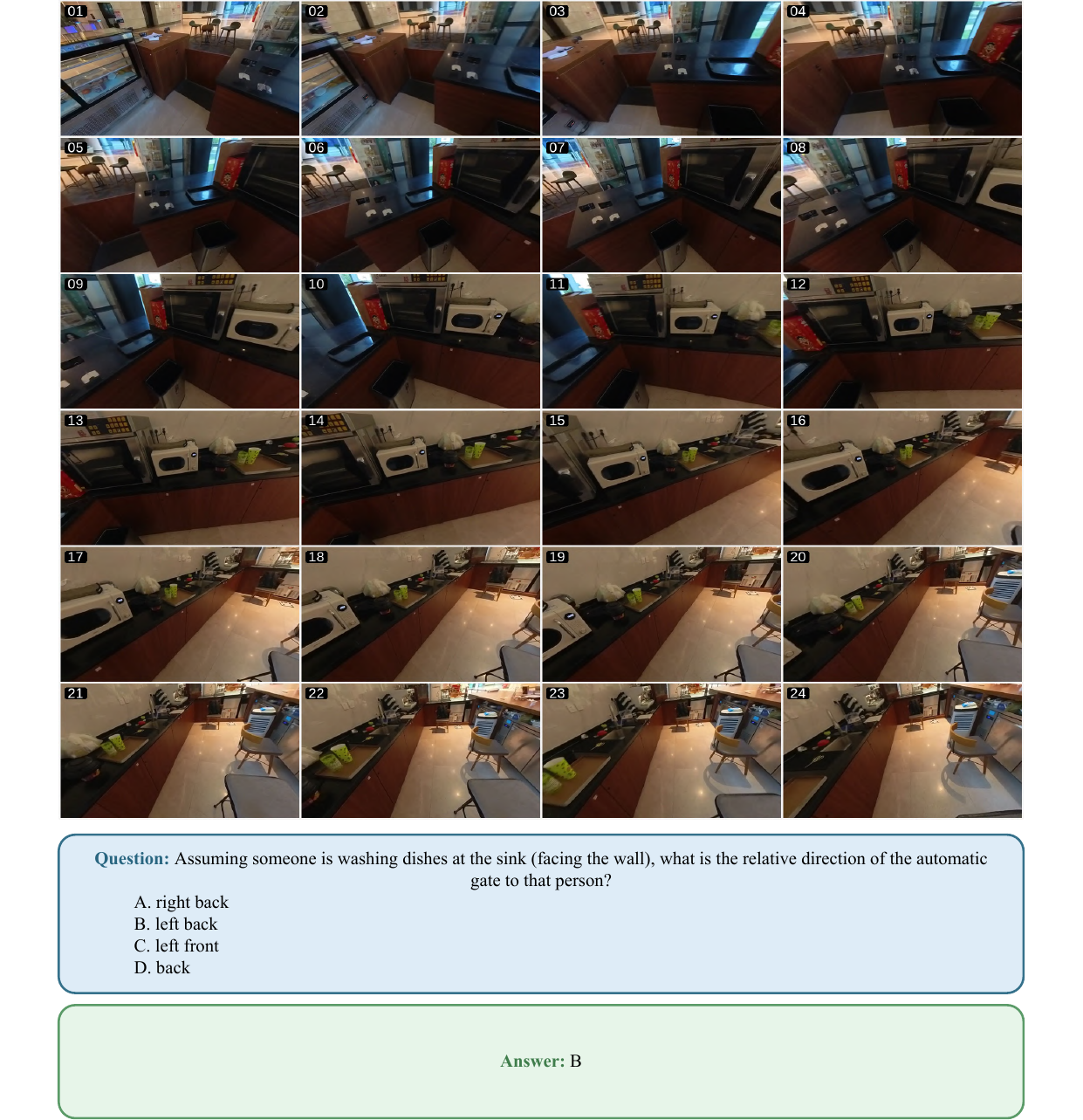}

    \caption{An example of spatial imagination and orientation reasoning a scene with multi-view observations.}
    \label{fig:multi-image-spatial-orientation}
\end{figure*}

\begin{figure*}[p]
    \centering
    \includegraphics[
        width=\textwidth,
        height=0.86\textheight,
        keepaspectratio
    ]{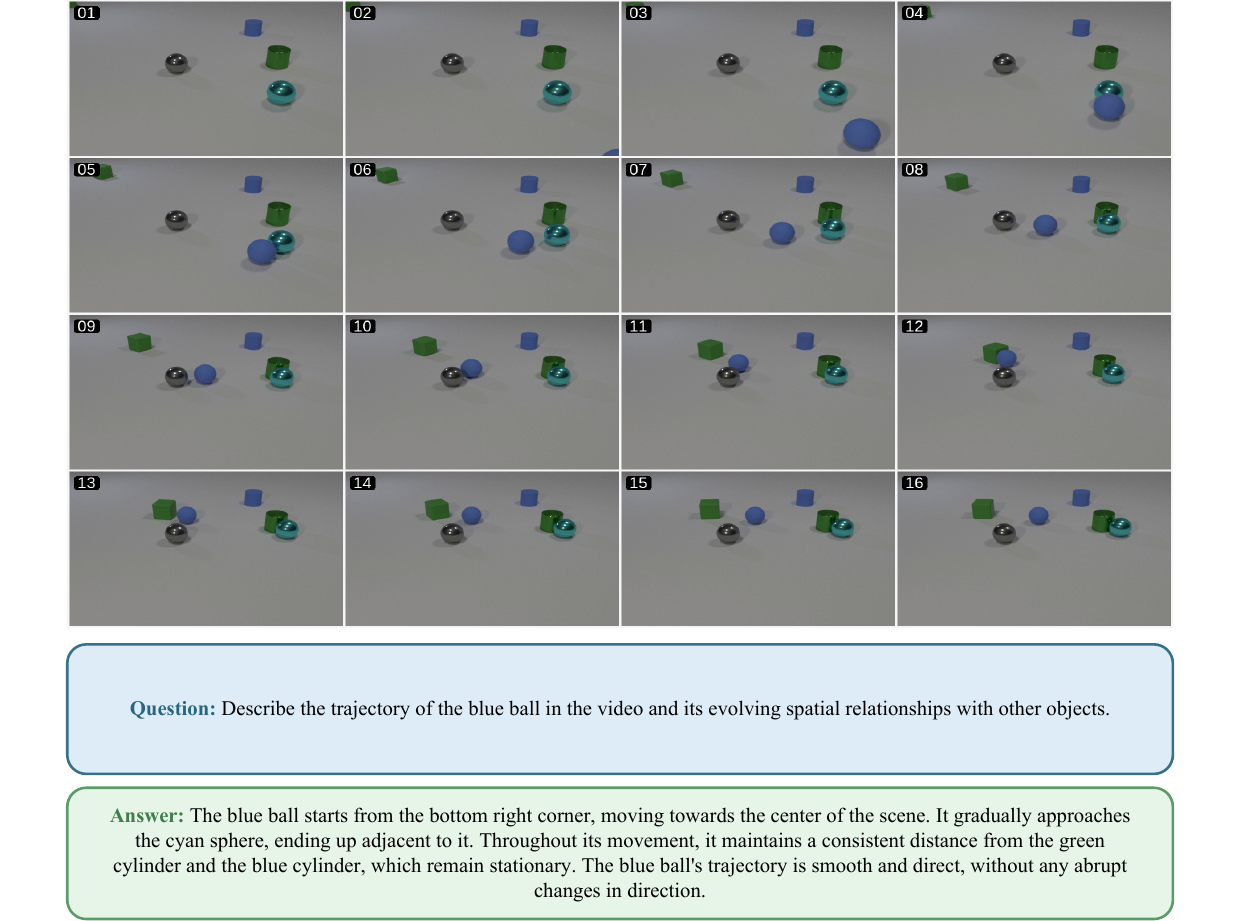}

    \caption{An example of object motion trajectory understanding in a short video.}
    \label{fig:multi-image-video-understanding}
\end{figure*}

\begin{figure*}[p]
    \centering
    \includegraphics[
        width=\textwidth,
        height=0.86\textheight,
        keepaspectratio
    ]{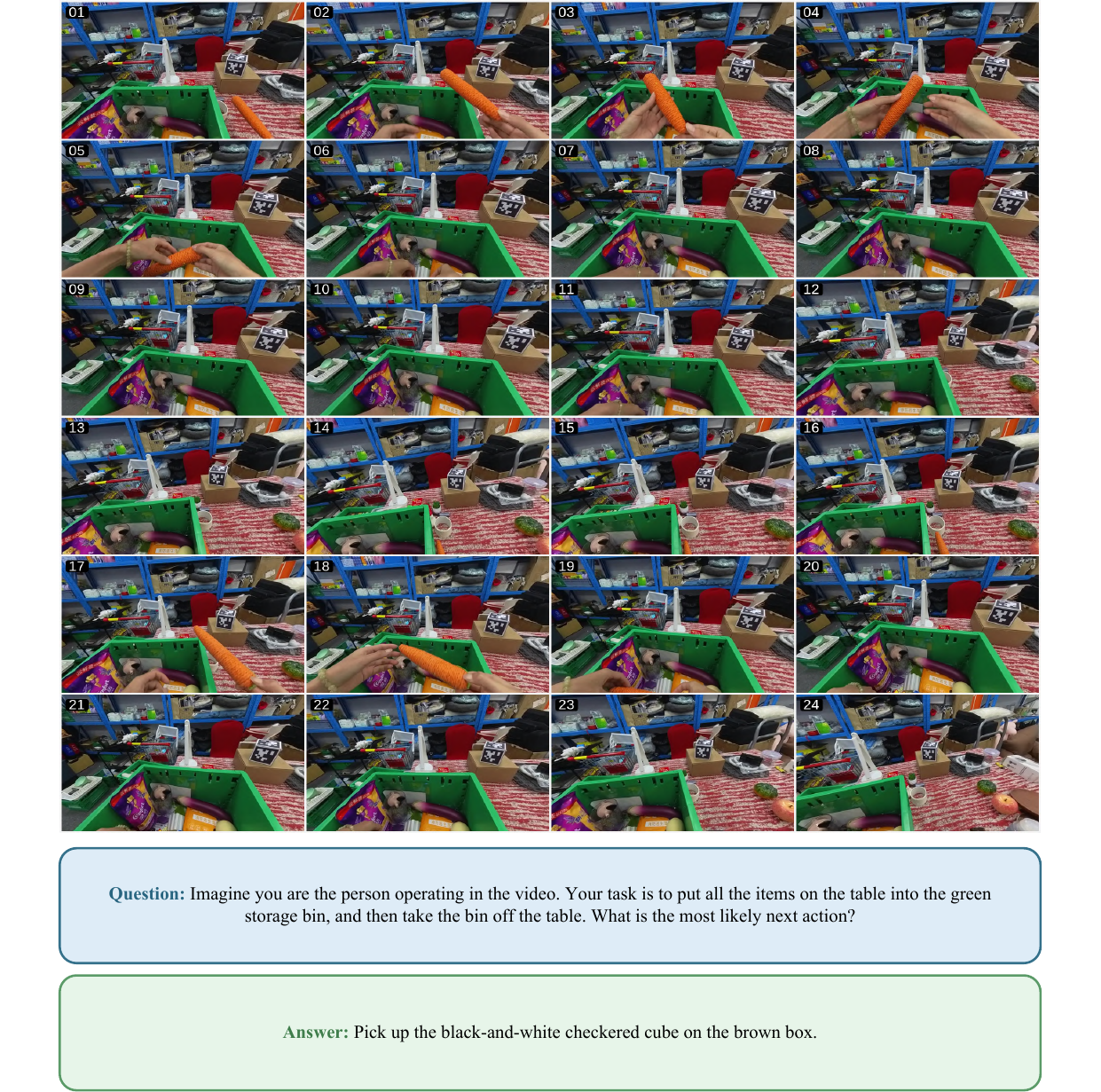}

    \caption{An example of future prediction for a video.}
    \label{fig:multi-image-future-prediction}
\end{figure*}

This appendix presents qualitative visualizations of PhysBrain 1.5 across three task categories: pointing~(Figure~\ref{fig:point-examples-vis}), trajectory point generation~(Figure~\ref{fig:trajectory-examples-vis}), and spatial relationship Q\&A~(Figures~\ref{fig:spatial-qa-examples-vis}, \ref{fig:multi-image-scale-estimation}, \ref{fig:multi-image-spatial-orientation}, \ref{fig:multi-image-video-understanding}, and \ref{fig:multi-image-future-prediction}). These examples are drawn from sources outside the 28 embodied benchmarks used in our evaluation, illustrating the ability of our unified embodied foundation model to generalize to a broader range of scenarios.

\clearpage

\clearpage

\clearpage

\clearpage

\end{document}